\documentclass[10pt,twocolumn,letterpaper]{article}

\usepackage[pagenumbers]{wacv} %

\usepackage[dvipsnames]{xcolor}
\usepackage{makecell}
\usepackage{sidecap}

\newcommand\edit[1]{#1}

\newcommand\revise[1]{#1}

\newcommand{\ourmethod}[1]{\textit{Learn-to-Steer}}
\newcommand{\spatialrelations}[1]{\textit{\{"above", "below", "to the left of", "to the right of"\}}}

\definecolor{wacvblue}{rgb}{0.21,0.49,0.74}
\usepackage[pagebackref,breaklinks,colorlinks,allcolors=wacvblue]{hyperref}

\def\wacvPaperID{872}  %
\def\confName{WACV}
\def\confYear{2026}

\title{Data-Driven Loss Functions for Inference-Time Optimization in Text-to-Image}

\author{
Sapir Esther Yiflach$^{1}$ \quad
Yuval Atzmon$^{2}$ \quad
Gal Chechik$^{1,2}$ \\[0.25em]
$^{1}$Bar-Ilan University \quad
$^{2}$NVIDIA \\[0.5em]
\texttt{Project page:~\url{https://learn-to-steer-paper.github.io}}
}

\begin{document}
\maketitle

\begin{abstract} 
\noindent
Text-to-image diffusion models can generate stunning visuals, yet they often fail at tasks children find trivial—like placing a dog to the right of a teddy bear rather than to the left. When combinations get more unusual—a giraffe above an airplane—these failures become even more pronounced. Existing methods attempt to fix these spatial reasoning failures through model fine-tuning or test-time optimization with handcrafted losses that are suboptimal. Rather than imposing our assumptions about spatial encoding, we propose learning these objectives directly from the model's internal representations.

We introduce \ourmethod{}, a novel framework that learns data-driven objectives for test-time optimization rather than handcrafting them. Our key insight is to train a lightweight classifier that decodes spatial relationships from the diffusion model's cross-attention maps, then deploy this classifier as a learned loss function during inference. Training such classifiers poses a surprising challenge: they can take shortcuts by detecting linguistic traces \revise{in the cross-attention maps, rather than learning true spatial patterns. We solve this by augmenting our training data with samples generated using prompts with incorrect relation words, which encourages the classifier to avoid linguistic shortcuts and learn spatial patterns from the attention maps.} Our method dramatically improves spatial accuracy: from 20\% to 61\% on FLUX.1-dev and from 7\% to 54\% on SD2.1 across standard benchmarks. \revise{It also generalizes to multiple relations with significantly improved accuracy.}
\end{abstract}

\begin{figure}[!htbp]
  \centering
    \includegraphics[width=0.9\linewidth, trim={0.cm 5.5cm 14.59cm 0cm},clip]{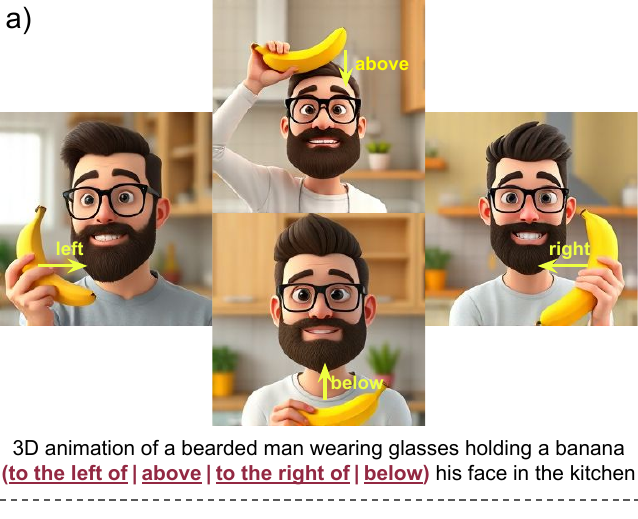} \\[-3pt] %
    \includegraphics[width=0.9\linewidth, trim={0.cm 2.45cm 14.15cm 0cm},clip]{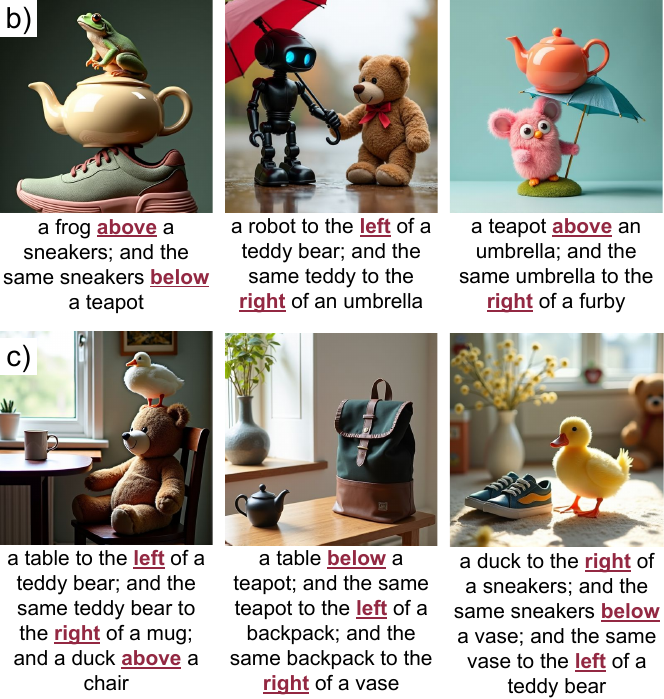} \\[-5pt] %
   \caption{\textbf{\ourmethod{}} learns how spatial relationships are encoded in attention maps to guide generation. \textbf{a)} Correctly renders all four orientations (above/below/left/right) ~\textbf{b, \revise{c})} Handles complex scenes with \textit{multiple spatial relationships} \revise{of up to five objects and three relations}. Prompts are for illustration purpose, see ~\cref{sec:multi} for actual prompt structure.   }
   \label{fig1}
   \vspace{-15pt}
\end{figure}

\section{Introduction}
\label{sec:intro}

Imagine asking a text-to-image model to generate ``a dog to the right of a teddy bear." What could be simpler? Yet surprisingly often, you might get a dog to the left, just a teddy bear with no dog at all, or perhaps a chimeric hybrid of them both. When we venture into more unusual territory like ``a cow above a boy" the situation only gets worse. Despite significant progress in text-to-image generation, where models like FLUX.1-dev~\cite{flux} can create high-quality images from text descriptions, they still struggle with understanding and depicting spatial relationships between objects.

Current approaches to improving text-image alignment generally fall into two camps. Some methods fine-tune the generative model ~\cite{han2025progressive, zhang2024compass, chatterjee2024getting} or train adapter layers~\cite{li2023gligen, chen2023reason}, risking degradation of the model's general capabilities. A more flexible paradigm is test-time optimization, which steers the generation process by optimizing the model's internal latent codes using a loss function. Current methods ~\cite{chefer2023attend, agarwal2023star, rassin2023linguistic, li2023divide, meral2024conform, 2025storm} typically rely on handcrafted loss functions applied to the model's attention maps. But here's the catch: these loss functions rely on simple heuristics that may be suboptimal  and address only narrow failure modes, like comparing centers of mass between object attention maps. 
This raises a fundamental question: instead of imposing our assumptions about how spatial relations should be encoded in attention maps, what if we could learn directly from the model's own internal representations how to steer it toward better alignment?

Spatial relations represent one type of text-to-image misalignment within a broader family of failures, including incorrect attribute binding, entity neglect, and object counting errors. In this work, we focus specifically on 2D spatial relationships (\eg, "above", "below", "left of", "right of") and propose a novel framework, \ourmethod{}, that learns loss functions for test-time optimization. The core of our method is to treat the steering problem as a learning problem. In an offline first step, we train a lightweight classifier that learns to decode spatial relationships directly from the cross-attention maps produced by the diffusion model. This classifier then serves as a data-driven loss function during inference, providing gradients to steer the latent code toward faithful representation of spatial relations. %
By learning rather than hand-crafting these guidance signals, we can leverage the complexity of how diffusion models encode spatial relationships across their internal representations, without manually specifying which patterns matter.

Our approach reveals a surprising challenge we term ``relation leakage": when naively trained, classifiers can achieve high accuracy by detecting linguistic traces of relation words in attention maps rather than actual spatial patterns. This creates a unique paradox: we need to predict ``left of'' while \revise{ignoring} textual traces of ``left of''--essentially learning to extract the visual manifestation of spatial concepts while ignoring their linguistic fingerprints. We address this through a simple yet effective data augmentation strategy that \revise{encourages the classifier to ignore these linguistic artifacts.}

Overall, using learned losses for steering generation significantly improves spatial relation accuracy across multiple diffusion models—from 7\% to 54\% on SD2.1 and from 20\% to 61\% on FLUX.1-dev -- while requiring no model fine-tuning, and avoiding the degradation often seen with fine-tuning approaches. Moreover, to the best of our knowledge, we are the first to demonstrate support for multiple simultaneous spatial relations, directly from input prompts. As shown in~\cref{fig1}, it can handle prompts with \revise{three} relations among \revise{five} objects. \revise{We quantify multiple relations generation accuracy using an extended evaluation scheme.}

Our main contributions are: (1) We show how the loss function for test-time steering can be learned from data rather than be handcrafted. (2)  We identify and solve the ``relation leakage'' problem, enabling effective training on cross-attention maps.
(3)  We demonstrate significant improvements over handcrafted losses, on standard benchmarks (GenEval and T2I-CompBench) across four different diffusion models without any model fine-tuning. (4) We handle multiple simultaneous spatial relations in a single image, demonstrating generalization beyond training on single relations.
\revise{(5) We introduce an extended evaluation scheme for quantifying multiple relation generation.}

\section{Related Work}
\label{sec:related_work}

Approaches to improve text-image alignment largely fall into two categories: those that modify the generative model itself, and those that guide it during inference.

The first strategy involves fine-tuning the model's weights. Some methods fine-tune the generator, for instance by using curriculum-based contrastive learning (EVOGEN~\cite{han2025progressive}), injecting token-ordering information with a curated spatially-aware dataset (COMPASS~\cite{zhang2024compass}), re-captioning datasets with spatial phrases (SPRIGHT~\cite{chatterjee2024getting}), or leveraging chain-of-thought with multimodal LLMs (Image-Generation CoT~\cite{zhang2025let}). Others optimize prompt embeddings directly (RRNet~\cite{wu2024rrnet}). A related approach involves training lighter adapter modules that inject new information, such as bounding boxes (GLIGEN~\cite{li2023gligen}), LLM-generated layouts (Reason-out-Your-Layout~\cite{chen2023reason}), or conditioning maps like Canny edges \cite{canny1986} (ControlNet~\cite{zhang2023controlnet}).
\edit{Similarly, layout-guided methods~\cite{chen2025ragd, zhou2024migc, zhou2025dis} use explicit spatial layouts of objects as additional input signals to spatially constrain the denoising process.}
In contrast, our work keeps the original pretrained model frozen \edit{and operates directly on text without requiring explicit layout inputs.}

The second major paradigm our work builds upon, is test-time optimization. Here, the latent code is steered during the denoising process using a guidance loss. However, the dominant approach is to rely on \textit{handcrafted} loss functions based on heuristics about how attention maps should behave. This includes methods that impose manually designed constraints on self- or cross-attention maps~\cite{chefer2023attend, agarwal2023star, rassin2023linguistic, li2023divide, meral2024conform, 2025storm}, as well as those using explicit layouts as alignment targets~\cite{binyamin2025make, wu2023harnessing, chen2024training, phung2024grounded}. Some methods optimize only the initial noise $z_T$ rather than the full trajectory: InitNO~\cite{guo2024initno} uses handcrafted attention-based losses, while ReNO~\cite{eyring2024reno} leverages external reward from human preference models. NPNet~\cite{zhou2025golden} takes a different approach by training a network to predict initial noises from a dataset of noise-image pairs filtered by human preference models, focusing on overall image quality rather than specific compositional aspects like spatial relations. Our work presents a fundamental departure: instead of hand-engineering objectives or relying on external models, we \textit{learn} the function to optimize directly from the model's own internal representations, capturing the often complex, patterns of spatial encoding.

Other strategies leverage powerful, external VLMs or MLLMs to provide feedback, using their scores to refine or rank generated images~\cite{singh2023vqadivide, li2024genaievaluating, li2025reflect}. In contrast, we use a lightweight classifier trained on the generator's own internal signals, avoiding dependence on large, black-box models. %

\noindent\textbf{Multiple Spatial Relations.} Generating scenes with multiple spatial relations is a long-standing challenge. Recent benchmarks on simple geometric shapes confirm that even state-of-the-art text-to-image models struggle with this task~\cite{hong2024evaluating}. A common strategy is to generate an intermediate layout to guide the process. Early works used GNNs on scene graphs to produce the intermediate layout, and were often limited to simple synthetic shapes~\cite{herzig2020learning, bar2020compositional, ivgi2021scene, farshad2023scenegenie}. More recent layout-guided methods use LLMs to interpret free-form text, yet their focus is typically on object counts and attributes rather than multiple explicit spatial relations~\cite{sella2024instancegen, khan2024composeanything}. While compositional diffusion approaches can handle multiple relations, they have primarily been demonstrated on simple synthetic shapes~\cite{liu2022compositional}. In contrast, our method demonstrates this capability directly from a text prompt on realistic images, without requiring an explicit intermediate layout.

\section{Preliminaries: Test-time optimization in text-to-image models}
\label{sec:preliminaries}

While text-to-image diffusion models produce high-quality images, they often struggle with complex prompts or fail to represent all specified concepts. Test-time optimization addresses these limitations by dynamically steering the generation process toward desired objectives without requiring model retraining. The method works by intervening in the model's iterative denoising process, which generates an image by progressively removing noise from an initial random latent $z_T$ to the final latent representation of the image $z_0$.

At each timestep $t$ of this process, the model's transformer blocks produce a set of cross-attention maps $A_t \in \mathbb{R}^{h\times w\times N}$, where $h \times w$ is the spatial resolution and $N$ is the number of text tokens. Each map $A_t[:,:,i]$ indicates how token $i$ influences each spatial location, forming a bridge between words and image regions.

Test-time optimization uses these maps to compute a loss $\mathcal{L}(A_t)$ that quantifies alignment with the desired objective at selected timesteps $t \in \mathcal{T}$, then updates the latent via gradient descent: $z'_t = z_t - \alpha_t \nabla_{z_t} \mathcal{L}$. While any differentiable signal could be used, cross-attention maps have emerged as the primary optimization target due to their interpretability and direct connection to text-image alignment.

\section{Method}
\label{sec:method}

Text-to-image diffusion models excel at generating high-quality images, yet they often struggle with accurately representing relationships between objects. %
While prior work has addressed these alignment challenges through handcrafted loss functions, such approaches are inherently limited: they require domain expertise to design, may encode sub-optimal heuristics, and often target narrow failure modes. What if, instead of manually engineering these losses, we could learn them directly from the model's own internal representations?

We propose exactly that: a method that learns to interpret and guide diffusion models by understanding how relationships are encoded in their cross-attention maps. Our approach consists of two stages. First, we train a relation classifier that learns to decode spatial relationships from cross-attention patterns. Second, we employ this classifier as a learned loss function during test-time optimization, steering the generation process toward faithful representation of the prompted relationships. %

\begin{figure}[htbp]
    \setlength{\abovecaptionskip}{4pt}
    \setlength{\belowcaptionskip}{0pt}
\centering
\includegraphics[width=\linewidth,trim={0.cm 0cm 0cm 0.cm},clip]{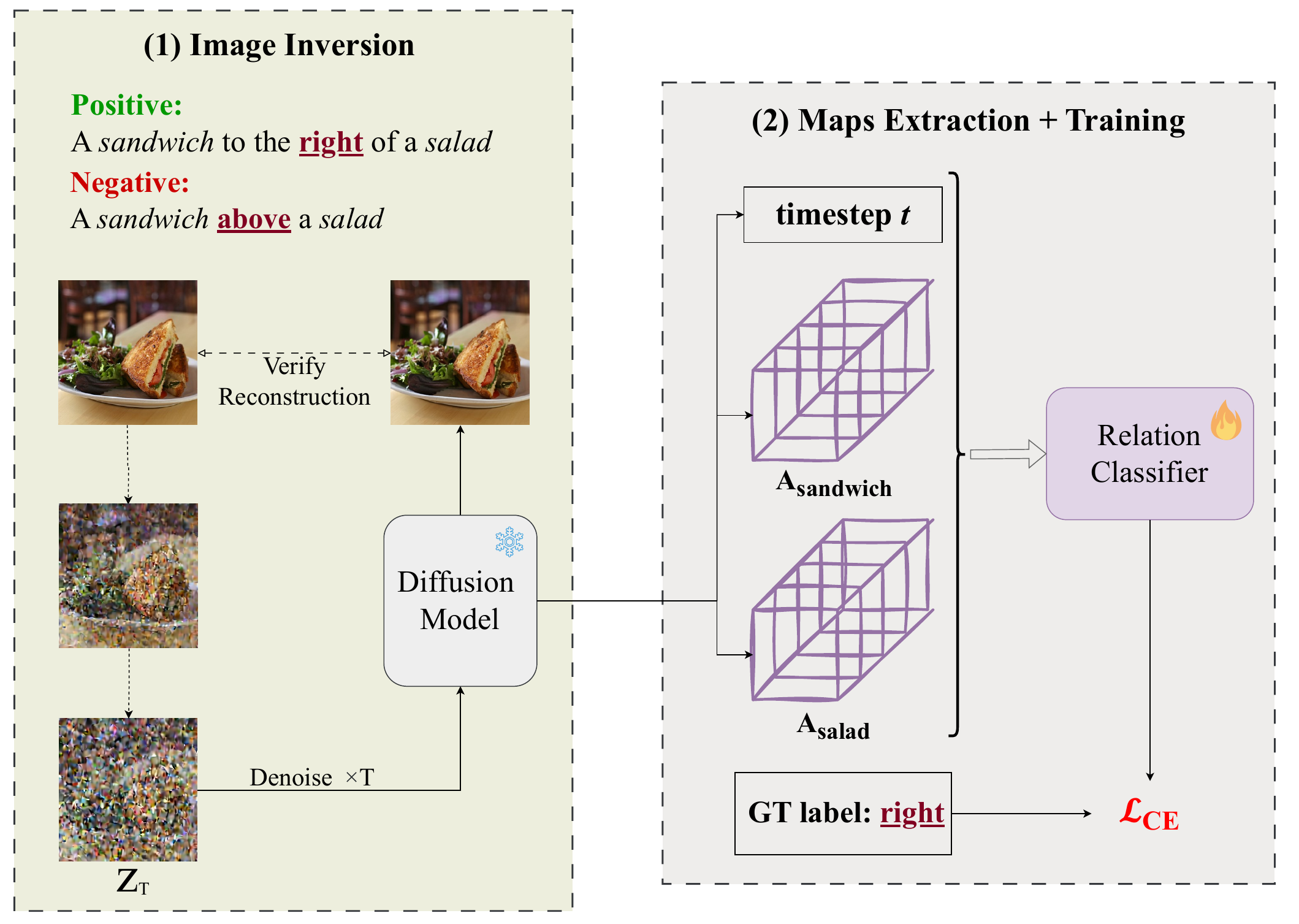}
    \caption{
     \textbf{Classifier training pipeline.}
     Given a spatially-aligned image, we  \revise{augment the data with both correct and incorrect relation prompts} to prevent relation leakage. During denoising, we extract relevant attention maps to train our classifier.}
    \label{fig_training_pipeline}
\vspace{-5pt}
\end{figure}

\begin{figure}[htbp]
    \centering
    \includegraphics[width=\linewidth,trim={0.4cm 0cm 0cm 0.6cm},clip]{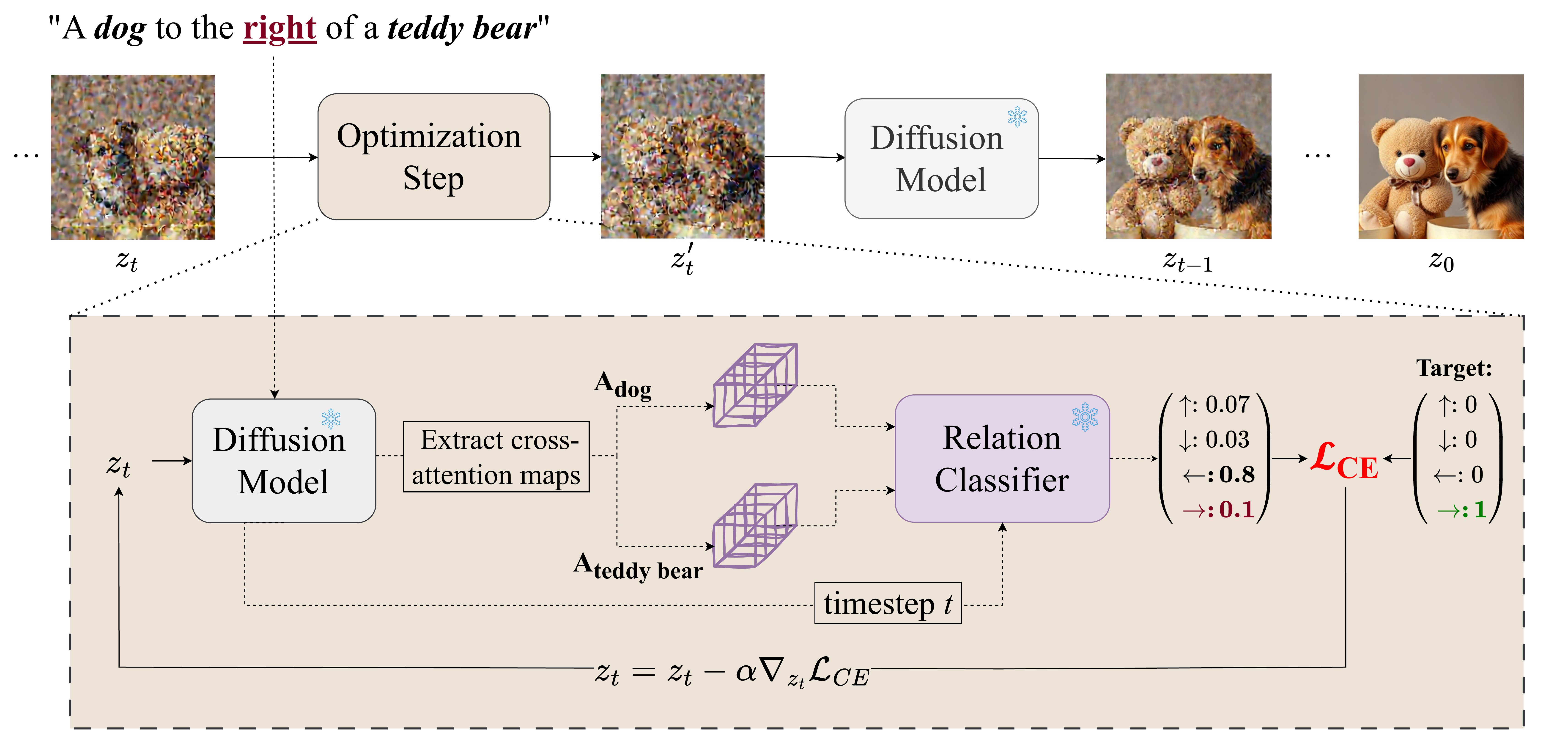}
    \caption{ \textbf{Test-time optimization pipeline.} During inference, we extract the relevant cross-attention maps when denoising $z_t$ and evaluate their relationship using our trained relation classifier. We then update the latent noise with backpropagation.}
    \label{fig_inference_pipeline}
\vspace{-5pt}
\end{figure}

\vspace{-5pt}
\subsection{Learning to Decode Relations from Attention}
\label{sec:method_classifier}

The foundation of our approach is a classifier $\mathcal{C}$, taking cross-attention maps from two objects and predicts their spatial relation. Training such a classifier reveals surprising properties of how diffusion models encode information.

~\newline
\noindent\textbf{The Relation Leakage Problem.}
\revise{Consider training a classifier on attention maps from inverted images. Given an image of ``a dog to the left of a cat'', we could invert it using this prompt, extract attention maps, and label them as ``left of''. However, such classifiers achieved high training accuracy but failed during generation. We discovered the cause: instead of analyzing spatial patterns, the classifier detected subtle traces of the relation word (``left'') encoded within the maps. This \textit{relation leakage} meant the classifier recognized linguistic artifacts rather than visual evidence. For example, when we inverted images using incorrect prompts, like describing a dog actually left of a cat as ``above''  - the classifier predicted ``above'' rather than the true ``left of'', confirming it was reading the prompted relation from the attention maps rather than the actual spatial configuration.}

~\newline
\noindent\textbf{Addressing relation leakage.}
We aim to make the relation classifier learn representations that do not depend on  linguistic cues from the relation word. This presents a unique challenge: unlike typical invariance learning where we ignore a protected attribute while predicting something else ~\citep{ganin2016domain, zemel2013learning, atzmon2020causal}, here the ``protected attribute'' (relation word traces) and our prediction target (spatial relation) are semantically same. We should predict ``left'' while ignoring linguistic traces of ``left'' in the attention maps.

\revise{To reduce leakage, we introduce an augmentation strategy to make the classifier ignore the linguistic cues coming from the inversion prompt. For each training image, we perform inversion twice: once with a \textit{``mismatching''} prompt that correctly describes the spatial relation (e.g., ``A dog to the left of a cat''), and once with a prompt containing a randomly sampled incorrect relation (e.g., ``A dog above a cat''). Crucially, both sets of extracted attention maps are labeled with the ground-truth relation observed in the image (``left of'' in this example). This ensures that both prompts generate positive training samples for the classifier. See ~\cref{dual_inv_supp} for potential limitations.} %

This augmentation strategy solves the leakage problem: since the same spatial configuration now appears with different relation words in the prompts, the classifier cannot rely on linguistic shortcuts. It must learn to rely on spatial patterns rather than textual artifacts, thereby avoiding relation leakage.

The training process is illustrated in~\cref{fig_training_pipeline}, while the full test-time optimization pipeline is shown in~\cref{fig_inference_pipeline}.
The detailed architecture of our relation classifier is described in Appendix~\ref{sec:architecture}.

\noindent\textbf{Training Data: Real and Synthetic Images.}
Our training dataset consists of (1) real images from GQA~\citep{hudson2019gqa} and (2) synthetically generated images from Image-Generation-CoT\cite{zhang2025let}. The real images provide naturalistic complexity and diverse spatial configurations but yield noisy attention maps when many objects are present. The synthetic images contain simpler scenes with fewer objects, producing cleaner attention maps that resemble and better align with the images encountered during generation.

\subsection{Test-Time Optimization with Learned Steering}
\label{sec:method_tto}
During generation, the classifier serves as a learned loss function. Given a prompt, we parse a single relation triplet for subject, relation and object $(s,r,o)$; and at timestep $t$, aggregate corresponding attention maps to obtain $\mathbf{A}_s$ and $\mathbf{A}_o$. The classifier predicts $\hat{p}=\mathcal{C}(\mathbf{A}_s,\mathbf{A}_o,t)$, where $\hat{p}$ is the predicted probability distribution over relation classes. We compute a cross-entropy loss against the desired relation $r$ from the prompt: $ \mathcal{L}_t=-\log \hat{p}[r]$, 
and update the latent code with its gradient: $z_t\leftarrow z_t-\alpha\nabla_{z_t}\mathcal{L}_t$. 
We optimize the first 50\% of the denoising steps. In FLUX.1-schnell, we only optimize the initial noise.

\subsection{Multiple Spatial Relations}
\label{sec:multi}
A key finding is that our loss, despite being trained only on single pairs of objects, generalizes effectively to prompts that specify multiple spatial relations -- ranging from two relations among three objects \revise{to more complex cases with 4-5 objects and up to three relations}. To accommodate such prompts, we alternate the optimization target at each denoising timestep. For each relation in the prompt, we perform a dedicated optimization step, updating the latent code for one relation before proceeding to the next.

Prompt phrasing has a noticeable impact on performance. While \cref{fig1}(b) shows structured prompts like ``a frog above a sneakers; and the \revise{same} sneakers below a teapot'' for illustrative purposes, we found that simpler, chained prompts of the form ``A relation B relation C'' are more effective and yield superior results. For example, ``a frog above a sneakers below a teapot''.

\subsection{Implementation Details}
\label{sec:implementation}

We describe here the main implementation details. Additional details are provided in Appendix~\ref{supp_impl_details}.

\noindent\textbf{Cross-Validation.}
All design choices and hyperparameters were selected through systematic validation on the downstream generation task. We made a validation set using the GenEval framework, carefully ensuring no overlap with test prompts. This set contains prompts with similar structure but different object combinations and spatial relations.

\noindent\textbf{Data.}
For Flux-based models, we use 20K real images from GQA and 4K synthetic images. For each image we extract the attention maps roughly every 5 denoising steps, and including the last (cleanest) step \eg $\{5, 10, 15, 20, 25, 49\}$ for multi-step models, and in FLUX.1-schnell we use $\{1, 3\}$. For SD-based models, we use a smaller set of training data, because the input dimension is lower ($7\times$ less maps in SD) -- with a balanced mixture of 4K real and 4K synthetic images. 
The GQA dataset is preprocessed to reduce ambiguity: we filter out images with multiple instances of the same object category and remove hypernym relations using WordNet~\citep{miller1995wordnet}.
We use inversion techniques to extract the attention maps of the training data for the denoising steps.
For Flux-based models, we use RF-Inversion~\cite{rout2025semantic}, for SD we use Fixed-Point Inversion~\cite{samuel2025gnri}, with a guidance-scale of $0$ for both inversion methods.

\begin{figure*}[htbp]
  \centering
    \includegraphics[width=0.975\linewidth]{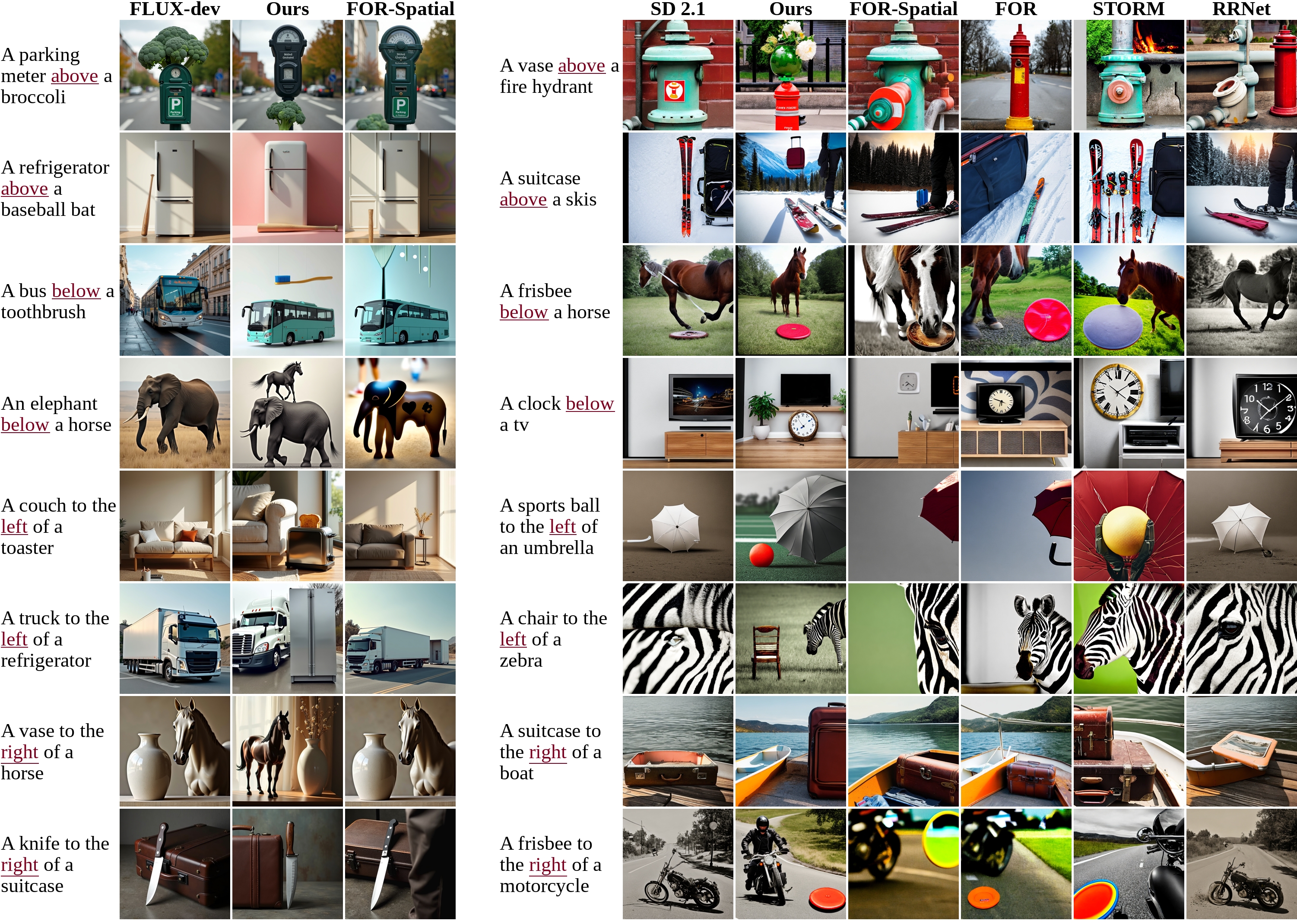}
   \caption{Comparison using FLUX.1-dev (left) and SD 2.1 (right) as base models with prompts from the GenEval~\cite{ghosh2023geneval} benchmark. For each prompt, the same seed is used for all methods. %
   }
   \label{fig:Fdev_sd2_geneval_r1}
   \vspace{-10pt}
\end{figure*}

\section{Experiments}
\label{sec:experiments}
We compare \ourmethod{} with strong baselines through quantitative and qualitative evaluation using two benchmark datasets and across \textit{four} text-to-image diffusion models: Flux.1-schnell~\cite{flux}, Flux.1-dev~\cite{flux}, SD2.1~\cite{SD2_SD14} and SD1.4~\cite{SD2_SD14}. Our method demonstrates significant improvements in spatial relations representation compared to baselines, which often suffer from entity neglect, remaining effective even for atypical scenarios like ``A cow above a boy". \revise{We then comprehensively evaluate generalization and scalability through out-of-distribution categories, compositional scaling to multiple relations, dense prompts, personalized models, and diagonal relations.} Finally, we study the properties of \ourmethod{} through an ablation study.

\subsection{Quantitative Analysis}
\begin{table}[ht]
  \centering
    \scalebox{0.85}
    {
    \renewcommand{\arraystretch}{0.9}
  \begin{tabular}{@{}lcc@{}}
    \toprule
    \textbf{Method} & \textbf{G. Pos.~\textuparrow} & \textbf{T. Spatial~\textuparrow} \\
    
    \midrule
    \textbf{FLUX.1-schnell} &  &  \\
    Vanilla FLUX.1-schnell  & 0.26 & 0.199 \\
    FOR-Spatial\dag~\cite{FOR_izadi2025fine} & {0.35} & {0.238} \\
    \ourmethod{} (Ours) & \textbf{0.52} & \textbf{0.309} \\
    
    \midrule
    \textbf{FLUX.1-dev} &  &  \\
    Vanilla FLUX.1-dev& 0.20 & 0.177 \\
    FOR-Spatial\dag~\cite{FOR_izadi2025fine} & 0.49 & {0.357} \\
    COMPASS~\cite{zhang2024compass} & {0.60} & - \\
    \revise{COMPASS$^{*}$} & \revise{0.59} & \revise{0.318} \\
    \ourmethod{} (Ours) & \textbf{0.61} & \textbf{0.392} \\
    \midrule
    \textbf{SD 2.1} &  &  \\
    Vanilla SD 2.1 & 0.07 & 0.089 \\
    FOR-Spatial\dag~\cite{FOR_izadi2025fine} & 0.27 & {0.211} \\
    FOR~\cite{FOR_izadi2025fine} & 0.34 & 0.192 \\
    STORM\dag~\cite{2025storm} & 0.19 & 0.111 \\
    RRNet~\cite{wu2024rrnet} & 0.09 & 0.106 \\
    COMPASS~\cite{zhang2024compass} & {0.51} & - \\
    \revise{COMPASS$^{*}$} & \revise{0.51} & \revise{0.257} \\
    \ourmethod{} (Ours) & \textbf{0.54} & \textbf{0.365} \\
    \midrule
    \textbf{SD 1.4} &  &  \\
    Vanilla SD 1.4 & 0.05 & 0.057 \\
    FOR-Spatial\dag~\cite{FOR_izadi2025fine} & 0.26 & {0.228} \\
    FOR~\cite{FOR_izadi2025fine} & 0.30 & 0.169 \\
    STORM\dag~\cite{2025storm} & 0.18 & 0.101 \\
    InitNO~\cite{guo2024initno} & 0.12 & 0.062 \\
    COMPASS~\cite{zhang2024compass} & \textbf{0.46} & - \\
    \revise{COMPASS$^{*}$} & \revise{0.43} & \revise{0.254} \\
    \ourmethod{} (Ours) & {0.36} & \textbf{0.278} \\
    
    \bottomrule
  \end{tabular}
  }
  \caption{Comparison of spatial relation scores from the GenEval (\textit{G. Pos.}) and T2I-CompBench (\textit{T. Spatial}) benchmarks grouped by base diffusion model. All results were evaluated using the same test set, except for COMPASS, which is reported from~\cite{zhang2024compass}. \revise{COMPASS$^{*}$ indicates evaluation using their provided code.} \dag~ indicates methods that use handcrafted spatial loss functions.
  }
  \label{tab:benchmark_results}
\end{table}

\begin{table}[ht]
  \centering
    \scalebox{0.92}
    {
    \renewcommand{\arraystretch}{0.9}
  \begin{tabular}{@{}lccc@{}}
    \toprule
    \textbf{Method} & \textbf{FLUX.1-dev} & \textbf{SD 2.1} & \textbf{SD 1.4} \\
    \midrule
    Vanilla & \revise{0.18} & \revise{0.15} & \revise{0.07} \\
    \revise{COMPASS$^{*}$} & \revise{0.58} & \revise{0.52} & \revise{0.40} \\
    \ourmethod{} (Ours) & \revise{\textbf{0.60}} & \revise{\textbf{0.60}} & \revise{\textbf{0.47}} \\
    \bottomrule
  \end{tabular}
  }
  \caption{\revise{Out-of-distribution (OOD) performance showing GenEval's (\textit{G. Pos.}) scores on categories that are out-of-distribution for our model and COMPASS.} 
  }
  \label{tab:ood_results}
  \vspace{-10pt}
\end{table}

\textbf{Baselines:}
We compare with several methods. \textbf{(1) COMPASS}~\cite{zhang2024compass} fine-tunes the model with spatially-aligned data and a text token ordering module. \textbf{(2) RRNet}~\cite{wu2024rrnet} optimizes prompt embeddings using spatially-aligned reference images. \textbf{(3) FOR}~\cite{FOR_izadi2025fine} employs a two-stage test-time approach: first optimizing latents $z_t$ with handcrafted losses, then refining initial noise $z_T$ using VLM feedback. \textbf{(4) FOR-Spatial}~\cite{FOR_izadi2025fine} uses only the handcrafted spatial loss component from FOR's first stage for test-time optimization--since FOR only supports SD1.4 and SD2.1, we adapt this component for Flux models by aggregating cross-attention maps per object (details in~\cref{supp_FOR_flux_impl}). \textbf{(5) STORM}~\cite{2025storm} uses handcrafted test-time losses based on optimal transport for spatial relations. \textbf{(6) InitNO}~\cite{guo2024initno} uses handcrafted attention-based loss functions for test-time optimization, focusing on entity neglect and attribute binding failures. We reproduce results for all methods using their shared code, except COMPASS code was not released--we compare with images from their paper in Appendix~\ref{supp_compass}, 
\edit{and in \ref{supp_layout_guided}, we further compare to layout-guided methods.}

\noindent\textbf{Benchmarks and Metrics}
We evaluate spatial alignment using two benchmarks, each with its own automated evaluation protocol.
\textbf{(1) GenEval}~\cite{ghosh2023geneval} provides 100 prompts covering four spatial relations: \{"above", "below", "left of", "right of"\}. We generate four images per prompt and report the percentage of spatially-accurate images among 400 total. Its evaluation metric (\textbf{G. Pos.}) verifies these by comparing the centroids of detected bounding boxes.
\textbf{(2) T2I-CompBench}~\cite{huang2023t2i} contains prompts for seven spatial relations: \{"on side of", "next to", "near", "on the left of", "on the right of", "on the bottom of", "on the top of"\}. We extract 216 prompts matching our four trained relations, generate ten images each, and report the percentage of spatially-accurate images. Its metric (\textbf{T. Spatial}) also uses bounding box centers, but applies stricter geometric rules involving the dominant axis of separation and low IoU.

\begin{table}[ht]
  \centering
    \scalebox{0.69}
    {
    \renewcommand{\arraystretch}{0.9}
  \begin{tabular}{@{}cc|ccc|ccc@{}}
    \toprule
    & & \multicolumn{3}{c|}{\revise{\large{\textbf{In-distribution categories}}}} & \multicolumn{3}{c}{\revise{\large{\textbf{OOD categories}}}} \\
    \cmidrule(lr){3-5} \cmidrule(lr){6-8}
    \revise{\textbf{Obj.}} & \revise{\textbf{Rel.}} & \revise{\textbf{Flux-dev}} & \revise{\textbf{Ours}} & \revise{\textbf{COMPASS$^{*}$}} & \revise{\textbf{Flux-dev}} & \revise{\textbf{Ours}} & \revise{\textbf{COMPASS$^{*}$}} \\

    \midrule
    \revise{\large 3} & \revise{\large 2} & \revise{\large 0.05} & \revise{\textbf{\large 0.36}} & \revise{\large 0.14} & \revise{\large 0.02} & \revise{\textbf{\large 0.28}} & \revise{\large 0.12} \\
    \revise{\large 4} & \revise{\large 2} & \revise{\large 0.04} & \revise{\textbf{\large 0.36}} & \revise{\large 0.21} & \revise{\large 0.04} & \revise{\textbf{\large 0.22}} & \revise{\large 0.14} \\
    \revise{\large 4} & \revise{\large 3} & \revise{\large 0.01} & \revise{\textbf{\large 0.20}} & \revise{\large 0.07} & \revise{\large 0.01} & \revise{\textbf{\large 0.19}} & \revise{\large 0.05} \\
    \revise{\large 5} & \revise{\large 3} & \revise{\large 0.01} & \revise{\textbf{\large 0.17}} & \revise{\large 0.05} & \revise{\large 0.01} & \revise{\textbf{\large 0.09}} & \revise{\large 0.05} \\

    \bottomrule
  \end{tabular}
  }
  \caption{\revise{\textbf{Multiple relations} accuracy across varying object and relation counts for in-distribution and OOD categories.} }
  \label{tab:multiple_relations}
  \vspace{-10pt}
\end{table}

\noindent\textbf{Quantitative Results:}
Table~\ref{tab:benchmark_results} reveals how learned objectives compare to handcrafted ones across four different models. On SD2.1, spatial accuracy rises from 0.07 to 0.54 on GenEval and from 0.089 to 0.365 on T2I-CompBench—transforming a model that barely works for spatial tasks into one that succeeds more than half the time. FLUX.1-dev shows similar patterns, improving from 0.20 to 0.61 and 0.177 to 0.392 respectively.
The results highlight two key findings. First, learned objectives consistently outperform handcrafted spatial losses: STORM achieves 0.19 on SD2.1 GenEval while our method reaches 0.54, and our approach shows similar advantages over FOR-Spatial across all tested models. Second, test-time optimization is competitive with state-of-the-art fine-tuning: COMPASS achieves 0.60 on FLUX.1-dev GenEval through model retraining, while our approach reaches 0.61 by steering the frozen model. Importantly, this pattern holds across architectures—from UNet-based SD models to MMDiT-based Flux variants. Notably, while COMPASS improves spatial relations, it degrades other capabilities—on SD2.1 color accuracy drops from 0.85 to 0.71 and counting accuracy falls from 0.44 to 0.20 (Appendix~\ref{supp_compass}). Our approach avoids these trade-offs by steering only when needed.

\noindent\textbf{Runtime:}
We analyze the computational costs of our approach. Our method involves an offline training phase followed by test-time optimization during inference. Classifier training requires 7 hours on an NVIDIA A100 80GB GPU. During inference, test-time optimization introduces computational overhead that varies by model: FLUX.1-schnell increases from 0.5s to 16.5s per image, FLUX.1-dev from 11s to 6 minutes, SD2.1 from 4.5s to 90s, and SD1.4 from 4.5s to 80s. While this overhead is substantial, it enables significant spatial accuracy improvements without requiring model fine-tuning. 
\revise{Finally, we analyzed runtime-accuracy tradeoffs for SD 2.1. Performing multiple optimization iterations (15 steps) only during the first 10\% of denoising steps while reducing to single optimization steps thereafter cuts runtime from 90s to 25s, while accuracy merely drops from 0.65 to 0.6 on the validation set. Early stages prove most critical - extending this regime to the first 20\% of steps (35s) restores accuracy to nearly 0.65.}

\subsection{Qualitative Analysis}
Figure~\ref{fig:Fdev_sd2_geneval_r1} \revise{(also Suppl Fig.~\ref{fig:unified_geneval})} show qualitative examples across four base models and several baselines. Compared to other methods, our approach consistently generates high-quality images that accurately reflect the spatial relations in the prompts. It also handles generating atypical and uncommon scenarios well, as can be seen in~\cref{fig:Fdev_sd2_geneval_r1} with \textit{``A bus below a toothbrush"} and in~\cref{fig:unified_geneval} with \textit{``An elephant below a surfboard}. 
In contrast, other baselines tend to neglect entities: in~\cref{fig:Fdev_sd2_geneval_r1}, none depict the vase from \textit{``A vase above a fire hydrant"}, and the zebra or the chair are missing from \textit{``A chair to the left of a zebra''}. These examples, along with~\cref{fig_common_base_failures} \revise{(Appendix)}, demonstrate how our method overcomes the three main failure modes: incorrect object placement, entity neglect, and object fusion. 

\begin{figure}
  \centering
    \includegraphics[width=0.8\linewidth, trim={0.cm 0.3cm 0cm 0.8cm},clip]{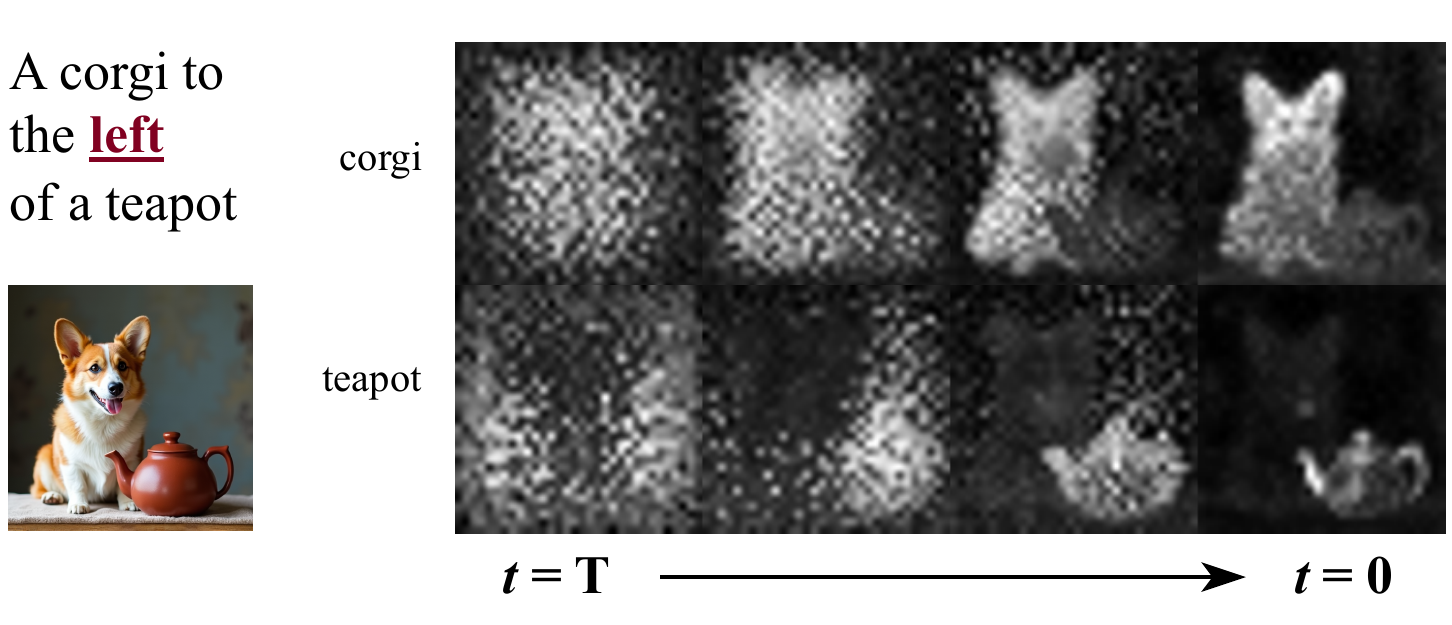}  %
   \caption{
   \revise{Evolution of cross-attention maps across denoising steps. Maps shown are after the steering step (See Fig. 3). At early steps ($t=T$), maps are diffuse and overlapping. During denoising, the corgi localizes on the left and the teapot on the right, converging to the intended spatial arrangement by $t=0$.}
   }
   \label{fig:cross_attn_fig}
   \vspace{-10pt}
\end{figure}

\textbf{Interpretability:}
\revise{
~\cref{fig:cross_attn_fig} visualizes cross-attention maps throughout denoising, showing how the model establishes spatial relations of two entities. Initially ($t=T$), attention maps are diffuse and overlapping. During denoising, they gradually separate - corgi localizing left, teapot right - matching the prompt. By $t=0$, objects are clearly disentangled with correct spatial alignment.  
}

\subsection{{Generalization and Scalability}}
\label{sec:generalization}

\revise{We evaluate our method's generalization capabilities across multiple dimensions: out-of-distribution categories, compositional complexity, dense prompts, personalized models, and diagonal spatial relations.}

\textbf{Generalization to unseen categories:}
\revise{
We test \ourmethod{} on object categories never seen during training. This out-of-distribution (OOD) evaluation uses 200 prompts across 16 object categories unseen by both our method and COMPASS~\cite{zhang2024compass}: Only 6 categories from T2I-CompBench  were external to both methods, and 10 additional categories were proposed by ChatGPT, ensuring no training overlap.
See \cref{supp_multi_relation_objects} for a full list. }
\revise{Table~\ref{tab:ood_results} shows that \ourmethod{} maintains strong performance on unseen object categories, achieving 0.60 on FLUX.1-dev and outperforming COMPASS on SD models (0.60 vs. 0.52 on SD2.1). }

\textbf{Compositional scaling to multiple relations:}
\revise{We evaluate how \ourmethod{} generalizes to prompts with multiple spatial relations -- a scenario never seen during training. We created 200 multi-relation prompts, ranging from two relations among three objects to complex cases with 4–5 objects and up to three relations, using both in-distribution (from the training set) and OOD categories as described above. We extended GenEval to support automatic verification of multiple relations and added open-vocabulary evaluation to recognize our new categories (details in \cref{supp_multi_relation}). Results are shown in Table~\ref{tab:multiple_relations}. FLUX.1-dev performs poorly (accuracy $\le$ 0.05); COMPASS, a training-based method, degrades severely with multiple relations (accuracy $\le$ 0.07 with 3 relations), 
despite strong single-relation performance. This sharp decline reflects the limitations of training-based approaches outside their training distribution. In contrast, \textbf{without any new training}, \ourmethod{} achieves substantial gains: for in-distribution categories with 2 relations, its accuracy of 0.36 on 3-4 objects matches the expected independent probability $(0.61)^2 = 0.372$.
Accuracy remains strong for OOD categories. 
These results highlight that our test-time optimization scales better with the number of relations, %
as performance degrades gracefully, even for OOD categories. See \cref{fig:supp_multi_rel_qual} (Appendix) for  qualitative comparisons.}

\textbf{Beyond four basic spatial relations:}
\revise{
To evaluate generality, we further trained a classifier for diagonal spatial relations (top-left, bottom-left, top-right, bottom-right). Since GenEval lacks an automated verifier for these relations, we could not use it to generate synthetic data for training the classifier, and therefore trained using only GQA data (see~\cref{diag_supp}). The classifier achieved similar performance (38\%) to the setting of non-diagonal relations that were trained on GQA alone (\cref{tab:ablation}, ``Real images only (GQA)''). See qualitative examples in~\cref{fig:diag_qual_fig} (Appendix).
}

\begin{figure}
  \centering

    \includegraphics[width=0.9\linewidth]{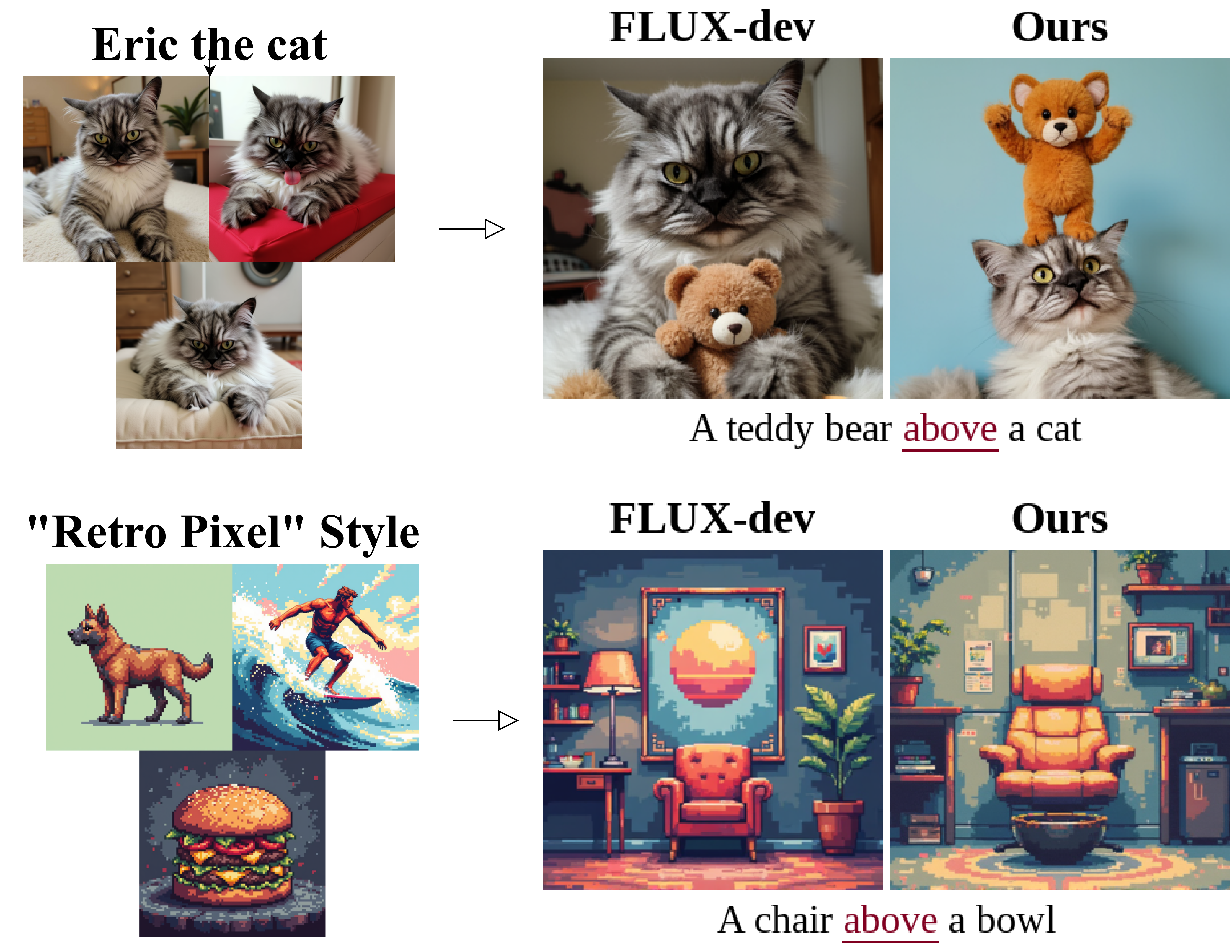}
    \caption{
    \revise{\textbf{Zero-shot transfer to personalized subjects and style LoRAs from HuggingFace.} More results in~\cref{fig:personal_fig_supp} (Appendix)} } %

   \label{fig:personal_fig}
   \vspace{-10pt}
\end{figure}

\textbf{Zero-shot transfer to personalized models:}
\revise{
\ourmethod{} works out-of-the-box with personalized LoRA models - establishing the spatial relations jointly with the personalized content.
~\cref{fig:personal_fig}, \ref{fig:personal_fig_supp} show two subjects - \emph{Eric the cat}~\cite{eric_the_cat_lora}, \emph{Margot Robbie}~\cite{margot_robbie_lora}, and a style - \emph{Retro Pixel}~\cite{retro_pixel_lora}. 
}

\textbf{Dense Prompts:}
\revise{\cref{fig:dense_fig} (Appendix) shows that our approach maintains spatial correctness under dense prompts enhanced with attributes and scene details.}

In Appendix~\ref{supp_compass}, qualitative comparisons with COMPASS show we achieve equivalent spatial accuracy while maintaining style and aesthetics closer to the base model. 
Additional results in Appendix~\ref{supp_qualitative_results} include extensive comparisons on both GenEval (Figures~\ref{fig:schnell_geneval_r1}, \ref{fig:dev_geneval_r2}, \ref{fig:sd2_geneval_r2}, and \ref{fig:sd14_geneval_r2}) and T2I-CompBench (Figures~\ref{fig:dev_t2i_r1}, \ref{fig:schnell_t2i_r1}, \ref{fig:sd2_t2i_r1}, and \ref{fig:sd14_t2i_r1}).

\subsection{Ablation Study}
\label{sec:ablations}

\begin{table}[ht]
    \centering
    \scalebox{0.85}
    {
    \renewcommand{\arraystretch}{0.9}
    \begin{tabular}{llc}
        \toprule
        \textbf{Component} & \textbf{Variant} & \textbf{G. Pos.~\textuparrow} \\
        \midrule
        \ourmethod{} (Ours) &  & 0.57 \\
        \midrule
        {Leakage Handling} & w/o \revise{leakage handling} & 0.425 \\
        \midrule
        {Training Data} & Real images only (GQA) & 0.365 \\
         & Synthetic images only & 0.373 \\
        \bottomrule
    \end{tabular}
    
    }
    \caption{Ablation results on FLUX.1-schnell validation set.}
    \label{tab:ablation}
\end{table}

We examine the key design choices of our method through systematic ablation on FLUX.1-schnell. Results are summarized in Table~\ref{tab:ablation}. 
\edit{See Appendix~\ref{sec:supp_attn_agg_ablation} for additional ablations.}

\textbf{Leakage Handling.} Our \revise{augmentation strategy, using mismatched prompts for inversion} is crucial for addressing relation leakage. Without it, training on positive prompts only yields 0.425. The \revise{mismatched-prompt} strategy eliminates this leakage, enabling our method to reach 0.57.

\textbf{Training Data Composition.} Real images from GQA alone achieve 0.365, while synthetic images alone reach 0.373. Real images provide naturalistic complexity but yield noisy attention maps in cluttered scenes. Synthetic images produce cleaner attention patterns that better match generation scenarios. Our full method combines both data sources, leveraging their complementary strengths.

\section{Limitations}
\label{sec:limitations}

\begin{figure}[t]
\centering
\includegraphics[width=0.9\linewidth,trim={0.cm 0cm 0.cm 0.7cm},clip]
{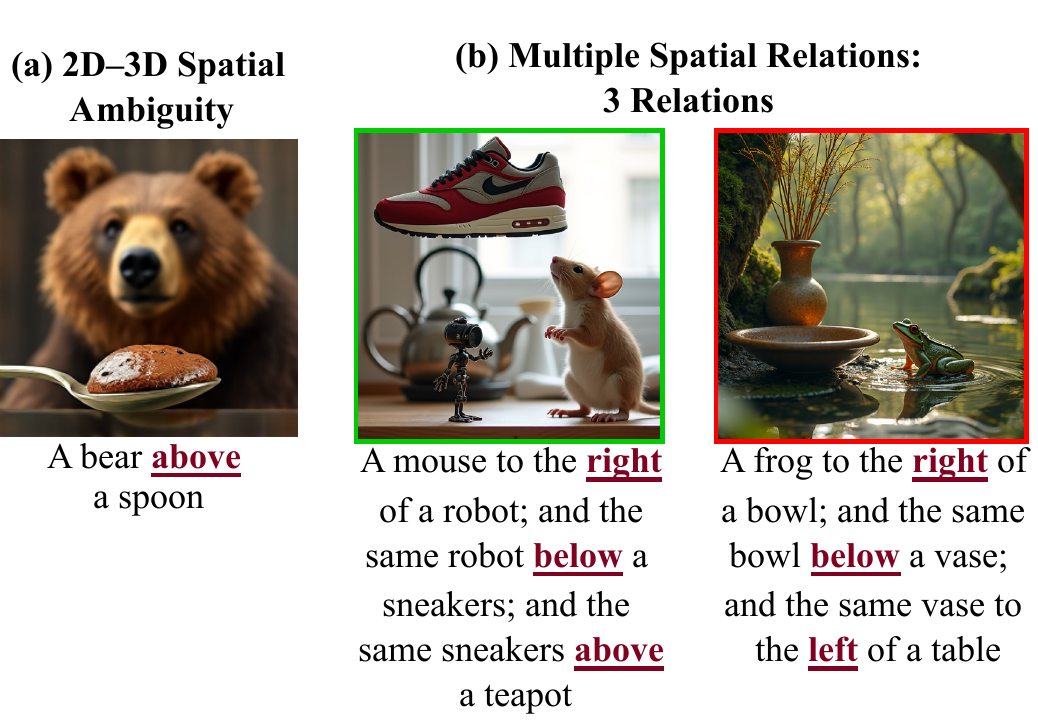} %
    \caption{
     \textbf{Method Limitations.}
     \textit{(a)} 2D spatial relations may not align with 3D scene geometry—the spoon appears below the bear in 2D but is in front of it in 3D.
     \textit{(b)} Multiple spatial relationships with \revise{three relations}: When prompted with four objects and \revise{three} spatial relations, our method can generate matching images. However, typical failure cases, like entity neglect still occur.
     }
    \label{fig_limitations}
\vspace{-10pt}
\end{figure}

Our method has several limitations. First, we train on 2D spatial relations, which can conflict with 3D scene understanding. As shown in~\cref{fig_limitations}(a), a spoon positioned in front of a bear in 3D space may appear "below" it in the 2D projection.
Second, handling scenarios with multiple spatial relationships becomes more challenging 
\revise{as the number of relations grow}.
~\cref{fig_limitations}(b) shows both successful generations and typical failure cases, such as entity neglect.

Finally, despite substantial improvements, there is still a lot of headroom for improvement. Entity neglect and incorrect positioning persist, as shown in the uncurated generation results in~\cref{fig:uncurated} in the Appendix.

\section{Conclusions}
\label{sec:conclusions}
We presented \ourmethod{}, a data-driven alternative to handcrafted spatial losses for test-time steering of text-to-image diffusion models. By training a lightweight classifier to decode spatial relationships from cross-attention maps and using it as a learned loss, we achieve substantial spatial accuracy improvements across multiple models without fine-tuning.
A key advantage of our approach is that it preserves the model's full capabilities by applying steering only when needed, avoiding the performance degradation often seen with fine-tuning methods. While test-time optimization increases inference time, this selective application ensures that the model's general generative abilities remain intact for prompts without spatial constraints. This cost enables applications requiring high spatial accuracy - including generation of multiple relations data.
Our work demonstrates that learning from the model's internal representations surpasses handcrafted objectives, opening new directions for improving text-to-image alignment.

\clearpage
\newpage

\section*{Acknowledgments}
\edit{This study was funded by the Israeli Ministry of Science, Israel-Singapore binational grant, and by a grant from the National Committee of Budgeting (VATAT). We thank Yoad Tewel, Ohad Rahamim, and Ben Fishman for providing feedback on an earlier version of this manuscript.}

{
    \small
    \bibliographystyle{ieeenat_fullname}
    \bibliography{main}

\begin{thebibliography}{59}
\providecommand{\natexlab}[1]{#1}
\providecommand{\url}[1]{\texttt{#1}}
\expandafter\ifx\csname urlstyle\endcsname\relax
  \providecommand{\doi}[1]{doi: #1}\else
  \providecommand{\doi}{doi: \begingroup \urlstyle{rm}\Url}\fi

\bibitem[Agarwal et~al.(2023)Agarwal, Karanam, Joseph, Saxena, Goswami, and Srinivasan]{agarwal2023star}
Aishwarya Agarwal, Srikrishna Karanam, KJ Joseph, Apoorv Saxena, Koustava Goswami, and Balaji~Vasan Srinivasan.
\newblock A-star: Test-time attention segregation and retention for text-to-image synthesis.
\newblock In \emph{Proceedings of the IEEE/CVF International Conference on Computer Vision}, pages 2283--2293, 2023.

\bibitem[Atzmon et~al.(2020)Atzmon, Kreuk, Shalit, and Chechik]{atzmon2020causal}
Yuval Atzmon, Felix Kreuk, Uri Shalit, and Gal Chechik.
\newblock A causal view of compositional zero-shot recognition.
\newblock In \emph{Advances in Neural Information Processing Systems (NeurIPS)}, 2020.

\bibitem[Bar et~al.(2021)Bar, Herzig, Wang, Rohrbach, Chechik, Darrell, and Globerson]{bar2020compositional}
Amir Bar, Roei Herzig, Xiaolong Wang, Anna Rohrbach, Gal Chechik, Trevor Darrell, and Amir Globerson.
\newblock Compositional video synthesis with action graphs.
\newblock In \emph{Proceedings of the 38th International Conference on Machine Learning}. PMLR, 2021.

\bibitem[Binyamin et~al.(2025)Binyamin, Tewel, Segev, Hirsch, Rassin, and Chechik]{binyamin2025make}
Lital Binyamin, Yoad Tewel, Hilit Segev, Eran Hirsch, Royi Rassin, and Gal Chechik.
\newblock Make it count: Text-to-image generation with an accurate number of objects.
\newblock In \emph{Proceedings of the Computer Vision and Pattern Recognition Conference}, pages 13242--13251, 2025.

\bibitem[Canny(1986)]{canny1986}
John Canny.
\newblock A computational approach to edge detection.
\newblock \emph{Pattern Analysis and Machine Intelligence, IEEE Transactions on}, 1986.

\bibitem[Chatterjee et~al.(2024)Chatterjee, Stan, Aflalo, Paul, Ghosh, Gokhale, Schmidt, Hajishirzi, Lal, Baral, et~al.]{chatterjee2024getting}
Agneet Chatterjee, Gabriela Ben~Melech Stan, Estelle Aflalo, Sayak Paul, Dhruba Ghosh, Tejas Gokhale, Ludwig Schmidt, Hannaneh Hajishirzi, Vasudev Lal, Chitta Baral, et~al.
\newblock Getting it right: Improving spatial consistency in text-to-image models.
\newblock In \emph{European Conference on Computer Vision}, pages 204--222. Springer, 2024.

\bibitem[Chefer et~al.(2023)Chefer, Alaluf, Vinker, Wolf, and Cohen-Or]{chefer2023attend}
Hila Chefer, Yuval Alaluf, Yael Vinker, Lior Wolf, and Daniel Cohen-Or.
\newblock Attend-and-excite: Attention-based semantic guidance for text-to-image diffusion models.
\newblock \emph{ACM transactions on Graphics (TOG)}, 42\penalty0 (4):\penalty0 1--10, 2023.

\bibitem[Chen et~al.(2024)Chen, Laina, and Vedaldi]{chen2024training}
Minghao Chen, Iro Laina, and Andrea Vedaldi.
\newblock Training-free layout control with cross-attention guidance.
\newblock In \emph{Proceedings of the IEEE/CVF winter conference on applications of computer vision}, pages 5343--5353, 2024.

\bibitem[Chen et~al.(2023)Chen, Liu, Yang, Yuan, You, Liu, and Yang]{chen2023reason}
Xiaohui Chen, Yongfei Liu, Yingxiang Yang, Jianbo Yuan, Quanzeng You, Li-Ping Liu, and Hongxia Yang.
\newblock Reason out your layout: Evoking the layout master from large language models for text-to-image synthesis.
\newblock \emph{arXiv preprint arXiv:2311.17126}, 2023.

\bibitem[Chen et~al.(2025)Chen, Li, Wang, Chen, Jiang, Li, Wang, Yang, and Tai]{chen2025ragd}
Zhennan Chen, Yajie Li, Haofan Wang, Zhibo Chen, Zhengkai Jiang, Jun Li, Qian Wang, Jian Yang, and Ying Tai.
\newblock Ragd: Regional-aware diffusion model for text-to-image generation.
\newblock In \emph{Proceedings of the IEEE/CVF International Conference on Computer Vision}, pages 19331--19341, 2025.

\bibitem[dewei Zhou et~al.(2025)dewei Zhou, Xie, Yang, and Yang]{zhou2025dis}
dewei Zhou, Ji Xie, Zongxin Yang, and Yi Yang.
\newblock 3{DIS}: Depth-driven decoupled image synthesis for universal multi-instance generation.
\newblock In \emph{The Thirteenth International Conference on Learning Representations}, 2025.

\bibitem[Eyring et~al.(2024)Eyring, Karthik, Roth, Dosovitskiy, and Akata]{eyring2024reno}
Luca Eyring, Shyamgopal Karthik, Karsten Roth, Alexey Dosovitskiy, and Zeynep Akata.
\newblock Reno: Enhancing one-step text-to-image models through reward-based noise optimization.
\newblock \emph{Advances in Neural Information Processing Systems}, 37:\penalty0 125487--125519, 2024.

\bibitem[Farshad et~al.(2023)Farshad, Yeganeh, Chi, Shen, Ommer, and Navab]{farshad2023scenegenie}
Azade Farshad, Yousef Yeganeh, Yu Chi, Chengzhi Shen, Bj{\"o}rn Ommer, and Nassir Navab.
\newblock Scenegenie: Scene graph guided diffusion models for image synthesis.
\newblock In \emph{Proceedings - 2023 IEEE/CVF International Conference on Computer Vision Workshops, ICCVW 2023}, 2023.

\bibitem[Ganin et~al.(2016)Ganin, Ustinova, Ajakan, Germain, Larochelle, Laviolette, March, and Lempitsky]{ganin2016domain}
Yaroslav Ganin, Evgeniya Ustinova, Hana Ajakan, Pascal Germain, Hugo Larochelle, Fran{\c{c}}ois Laviolette, Mario March, and Victor Lempitsky.
\newblock Domain-adversarial training of neural networks.
\newblock \emph{Journal of Machine Learning Research}, 2016.

\bibitem[Ghosh et~al.(2023)Ghosh, Hajishirzi, and Schmidt]{ghosh2023geneval}
Dhruba Ghosh, Hannaneh Hajishirzi, and Ludwig Schmidt.
\newblock Geneval: An object-focused framework for evaluating text-to-image alignment.
\newblock \emph{Advances in Neural Information Processing Systems}, 36:\penalty0 52132--52152, 2023.

\bibitem[ginipick(2024)]{eric_the_cat_lora}
ginipick.
\newblock Flux lora eric the cat.
\newblock \emph{URL https://huggingface.co/ginipick/flux-lora-eric-cat}, 2024.

\bibitem[Guo et~al.(2024)Guo, Liu, Cui, Li, Yang, and Huang]{guo2024initno}
Xiefan Guo, Jinlin Liu, Miaomiao Cui, Jiankai Li, Hongyu Yang, and Di Huang.
\newblock Initno: Boosting text-to-image diffusion models via initial noise optimization.
\newblock In \emph{Proceedings of the IEEE/CVF Conference on Computer Vision and Pattern Recognition}, pages 9380--9389, 2024.

\bibitem[Han et~al.(2025{\natexlab{a}})Han, Lee, Kim, Park, and Hwang]{2025storm}
Woojung Han, Yeonkyung Lee, Chanyoung Kim, Kwanghyun Park, and Seong~Jae Hwang.
\newblock Spatial transport optimization by repositioning attention map for training-free text-to-image synthesis.
\newblock In \emph{Proceedings of the IEEE/CVF Conference on Computer Vision and Pattern Recognition (CVPR)}, 2025{\natexlab{a}}.

\bibitem[Han et~al.(2025{\natexlab{b}})Han, Jin, Liu, and Liang]{han2025progressive}
Xu Han, Linghao Jin, Xiaofeng Liu, and Paul~Pu Liang.
\newblock Progressive compositionality in text-to-image generative models.
\newblock In \emph{The Thirteenth International Conference on Learning Representations}, 2025{\natexlab{b}}.

\bibitem[Herzig et~al.(2020)Herzig, Bar, Xu, Chechik, Darrell, and Globerson]{herzig2020learning}
Roei Herzig, Amir Bar, Huijuan Xu, Gal Chechik, Trevor Darrell, and Amir Globerson.
\newblock Learning canonical representations for scene graph to image generation.
\newblock In \emph{Computer Vision – ECCV 2020 - 16th European Conference, 2020, Proceedings}, 2020.

\bibitem[Honnibal et~al.(2020)Honnibal, Montani, Van~Landeghem, and Boyd]{Honnibal_spaCy_Industrial-strength_Natural_2020}
Matthew Honnibal, Ines Montani, Sofie Van~Landeghem, and Adriane Boyd.
\newblock {spaCy: Industrial-strength Natural Language Processing in Python}.
\newblock 2020.

\bibitem[Huang et~al.(2023)Huang, Sun, Xie, Li, and Liu]{huang2023t2i}
Kaiyi Huang, Kaiyue Sun, Enze Xie, Zhenguo Li, and Xihui Liu.
\newblock T2i-compbench: A comprehensive benchmark for open-world compositional text-to-image generation.
\newblock \emph{Advances in Neural Information Processing Systems}, 36:\penalty0 78723--78747, 2023.

\bibitem[Hudson and Manning(2019)]{hudson2019gqa}
Drew~A Hudson and Christopher~D Manning.
\newblock Gqa: A new dataset for real-world visual reasoning and compositional question answering.
\newblock In \emph{Proceedings of the IEEE/CVF conference on computer vision and pattern recognition}, pages 6700--6709, 2019.

\bibitem[{Hugging Face}(2025)]{huggingfaceT2I}
{Hugging Face}.
\newblock Text-to-image.
\newblock \emph{URL https://huggingface.co/docs/diffusers/en/using-diffusers/conditional\_image\_generation}, 2025.

\bibitem[Ivgi et~al.(2021)Ivgi, Benny, Ben-David, Berant, and Wolf]{ivgi2021scene}
Maor Ivgi, Yaniv Benny, Avichai Ben-David, Jonathan Berant, and Lior Wolf.
\newblock Scene graph to image generation with contextualized object layout refinement.
\newblock In \emph{2021 IEEE International Conference on Image Processing (ICIP)}, pages 2428--2432. IEEE, 2021.

\bibitem[Izadi et~al.(2025)Izadi, Hosseini, Tabar, Abdollahi, Saghafian, and Baghshah]{FOR_izadi2025fine}
Amir~Mohammad Izadi, Seyed Mohammad~Hadi Hosseini, Soroush~Vafaie Tabar, Ali Abdollahi, Armin Saghafian, and Mahdieh~Soleymani Baghshah.
\newblock Fine-grained alignment and noise refinement for compositional text-to-image generation.
\newblock \emph{arXiv preprint arXiv:2503.06506}, 2025.

\bibitem[Khan et~al.(2025)Khan, Chen, and Schmid]{khan2024composeanything}
Zeeshan Khan, Shizhe Chen, and Cordelia Schmid.
\newblock Composeanything: Composite object priors for text-to-image generation, 2025.

\bibitem[Labs(2024)]{flux}
Black~Forest Labs.
\newblock Flux.
\newblock \emph{URL https://blackforestlabs.ai/}, 2024.

\bibitem[Li et~al.(2024)Li, Lin, Pathak, Li, Fei, Wu, Xia, Zhang, Neubig, and Ramanan]{li2024genaievaluating}
Baiqi Li, Zhiqiu Lin, Deepak Pathak, Jiayao Li, Yixin Fei, Kewen Wu, Xide Xia, Pengchuan Zhang, Graham Neubig, and Deva Ramanan.
\newblock Evaluating and improving compositional text-to-visual generation.
\newblock In \emph{Proceedings of the IEEE/CVF Conference on Computer Vision and Pattern Recognition}, pages 5290--5301, 2024.

\bibitem[Li et~al.(2025)Li, Kallidromitis, Gokul, Koneru, Kato, Kozuka, and Grover]{li2025reflect}
Shufan Li, Konstantinos Kallidromitis, Akash Gokul, Arsh Koneru, Yusuke Kato, Kazuki Kozuka, and Aditya Grover.
\newblock Reflect-dit: Inference-time scaling for text-to-image diffusion transformers via in-context reflection.
\newblock \emph{arXiv preprint arXiv:2503.12271}, 2025.

\bibitem[Li et~al.(2023{\natexlab{a}})Li, Keuper, Zhang, and Khoreva]{li2023divide}
Yumeng Li, Margret Keuper, Dan Zhang, and Anna Khoreva.
\newblock Divide \& bind your attention for improved generative semantic nursing.
\newblock In \emph{34th British Machine Vision Conference 2023, {BMVC} 2023}, 2023{\natexlab{a}}.

\bibitem[Li et~al.(2023{\natexlab{b}})Li, Liu, Wu, Mu, Yang, Gao, Li, and Lee]{li2023gligen}
Yuheng Li, Haotian Liu, Qingyang Wu, Fangzhou Mu, Jianwei Yang, Jianfeng Gao, Chunyuan Li, and Yong~Jae Lee.
\newblock Gligen: Open-set grounded text-to-image generation.
\newblock In \emph{Proceedings of the IEEE/CVF conference on computer vision and pattern recognition}, pages 22511--22521, 2023{\natexlab{b}}.

\bibitem[Liu et~al.(2022)Liu, Li, Du, Torralba, and Tenenbaum]{liu2022compositional}
Nan Liu, Shuang Li, Yilun Du, Antonio Torralba, and Joshua~B Tenenbaum.
\newblock Compositional visual generation with composable diffusion models.
\newblock In \emph{Computer Vision--ECCV 2022: 17th European Conference, Tel Aviv, Israel, October 23--27, 2022, Proceedings, Part XVII}, pages 423--439. Springer, 2022.

\bibitem[Loshchilov and Hutter(2019)]{loshchilov2018decoupled}
Ilya Loshchilov and Frank Hutter.
\newblock Decoupled weight decay regularization.
\newblock In \emph{International Conference on Learning Representations}, 2019.

\bibitem[L\"uddecke and Ecker(2022)]{clipseg_lueddecke22_cvpr}
Timo L\"uddecke and Alexander Ecker.
\newblock Image segmentation using text and image prompts.
\newblock In \emph{Proceedings of the IEEE/CVF Conference on Computer Vision and Pattern Recognition (CVPR)}, pages 7086--7096, 2022.

\bibitem[Meral et~al.(2024)Meral, Simsar, Tombari, and Yanardag]{meral2024conform}
Tuna Han~Salih Meral, Enis Simsar, Federico Tombari, and Pinar Yanardag.
\newblock Conform: Contrast is all you need for high-fidelity text-to-image diffusion models.
\newblock In \emph{Proceedings of the IEEE/CVF Conference on Computer Vision and Pattern Recognition}, pages 9005--9014, 2024.

\bibitem[Miller(1995)]{miller1995wordnet}
George~A Miller.
\newblock Wordnet: a lexical database for english.
\newblock \emph{Communications of the ACM}, 38\penalty0 (11):\penalty0 39--41, 1995.

\bibitem[OpenAI(2025)]{chatgpt}
OpenAI.
\newblock Chatgpt.
\newblock \emph{URL https://chat.openai.com}, 2025.
\newblock Large language model.

\bibitem[Oquab et~al.(2023)Oquab, Darcet, Moutakanni, Vo, Szafraniec, Khalidov, Fernandez, Haziza, Massa, El-Nouby, Howes, Huang, Xu, Sharma, Li, Galuba, Rabbat, Assran, Ballas, Synnaeve, Misra, Jegou, Mairal, Labatut, Joulin, and Bojanowski]{oquab2023dinov2}
Maxime Oquab, Timothée Darcet, Theo Moutakanni, Huy~V. Vo, Marc Szafraniec, Vasil Khalidov, Pierre Fernandez, Daniel Haziza, Francisco Massa, Alaaeldin El-Nouby, Russell Howes, Po-Yao Huang, Hu Xu, Vasu Sharma, Shang-Wen Li, Wojciech Galuba, Mike Rabbat, Mido Assran, Nicolas Ballas, Gabriel Synnaeve, Ishan Misra, Herve Jegou, Julien Mairal, Patrick Labatut, Armand Joulin, and Piotr Bojanowski.
\newblock Dinov2: Learning robust visual features without supervision, 2023.

\bibitem[Phung et~al.(2024)Phung, Ge, and Huang]{phung2024grounded}
Quynh Phung, Songwei Ge, and Jia-Bin Huang.
\newblock Grounded text-to-image synthesis with attention refocusing.
\newblock In \emph{Proceedings of the IEEE/CVF Conference on Computer Vision and Pattern Recognition}, pages 7932--7942, 2024.

\bibitem[prithivMLmods(2024)]{retro_pixel_lora}
prithivMLmods.
\newblock Retro-pixel-flux-lora.
\newblock \emph{URL https://huggingface.co/prithivMLmods/Retro-Pixel-Flux-LoRA}, 2024.

\bibitem[punzel(2024)]{margot_robbie_lora}
punzel.
\newblock Flux lora - margot robbie.
\newblock \emph{URL https://huggingface.co/punzel/flux\_margot\_robbie}, 2024.

\bibitem[Radford et~al.(2021)Radford, Kim, Hallacy, Ramesh, Goh, Agarwal, Sastry, Askell, Mishkin, Clark, et~al.]{radford2021learningCLIP}
Alec Radford, Jong~Wook Kim, Chris Hallacy, Aditya Ramesh, Gabriel Goh, Sandhini Agarwal, Girish Sastry, Amanda Askell, Pamela Mishkin, Jack Clark, et~al.
\newblock Learning transferable visual models from natural language supervision.
\newblock In \emph{International conference on machine learning}, pages 8748--8763. PmLR, 2021.

\bibitem[Rassin et~al.(2023)Rassin, Hirsch, Glickman, Ravfogel, Goldberg, and Chechik]{rassin2023linguistic}
Royi Rassin, Eran Hirsch, Daniel Glickman, Shauli Ravfogel, Yoav Goldberg, and Gal Chechik.
\newblock Linguistic binding in diffusion models: Enhancing attribute correspondence through attention map alignment.
\newblock \emph{Advances in Neural Information Processing Systems}, 36:\penalty0 3536--3559, 2023.

\bibitem[Rombach et~al.(2022)Rombach, Blattmann, Lorenz, Esser, and Ommer]{SD2_SD14}
Robin Rombach, Andreas Blattmann, Dominik Lorenz, Patrick Esser, and Bj\"orn Ommer.
\newblock High-resolution image synthesis with latent diffusion models.
\newblock In \emph{Proceedings of the IEEE/CVF Conference on Computer Vision and Pattern Recognition (CVPR)}, pages 10684--10695, 2022.

\bibitem[Rout et~al.(2025)Rout, Chen, Ruiz, Caramanis, Shakkottai, and Chu]{rout2025semantic}
L Rout, Y Chen, N Ruiz, C Caramanis, S Shakkottai, and W Chu.
\newblock Semantic image inversion and editing using rectified stochastic differential equations.
\newblock In \emph{The Thirteenth International Conference on Learning Representations}, 2025.

\bibitem[Samuel et~al.(2025)Samuel, Meiri, Maron, Tewel, Darshan, Avidan, Chechik, and Ben-Ari]{samuel2025gnri}
Dvir Samuel, Barak Meiri, Haggai Maron, Yoad Tewel, Nir Darshan, Shai Avidan, Gal Chechik, and Rami Ben-Ari.
\newblock Lightning-fast image inversion and editing for text-to-image diffusion models.
\newblock In \emph{The Thirteenth International Conference on Learning Representations}, 2025.

\bibitem[Sella et~al.(2025)Sella, Kleiman, and Averbuch-Elor]{sella2024instancegen}
Etai Sella, Yanir Kleiman, and Hadar Averbuch-Elor.
\newblock Instancegen: Image generation with instance-level instructions, 2025.

\bibitem[Sim et~al.(2024)Sim, Lee, Tan, and Tan]{hong2024evaluating}
Shang~Hong Sim, Clarence Lee, Alvin Tan, and Cheston Tan.
\newblock Evaluating the generation of spatial relations in text and image generative models, 2024.

\bibitem[Singh and Zheng(2023)]{singh2023vqadivide}
Jaskirat Singh and Liang Zheng.
\newblock Divide, evaluate, and refine: Evaluating and improving text-to-image alignment with iterative vqa feedback.
\newblock \emph{Advances in Neural Information Processing Systems}, 36:\penalty0 70799--70811, 2023.

\bibitem[Vaswani et~al.(2017)Vaswani, Shazeer, Parmar, Uszkoreit, Jones, Gomez, Kaiser, and Polosukhin]{vaswani2017attention}
Ashish Vaswani, Noam Shazeer, Niki Parmar, Jakob Uszkoreit, Llion Jones, Aidan~N Gomez, \L~ukasz Kaiser, and Illia Polosukhin.
\newblock Attention is all you need.
\newblock In \emph{Advances in Neural Information Processing Systems}, 2017.

\bibitem[Wu et~al.(2023)Wu, Liu, Zhao, Bui, Lin, Zhang, and Chang]{wu2023harnessing}
Qiucheng Wu, Yujian Liu, Handong Zhao, Trung Bui, Zhe Lin, Yang Zhang, and Shiyu Chang.
\newblock Harnessing the spatial-temporal attention of diffusion models for high-fidelity text-to-image synthesis.
\newblock In \emph{Proceedings of the IEEE/CVF International Conference on Computer Vision}, pages 7766--7776, 2023.

\bibitem[Wu et~al.(2024)Wu, Yang, and Wang]{wu2024rrnet}
Yinwei Wu, Xingyi Yang, and Xinchao Wang.
\newblock Relation rectification in diffusion model.
\newblock In \emph{Proceedings of the IEEE/CVF Conference on Computer Vision and Pattern Recognition}, pages 7685--7694, 2024.

\bibitem[Zemel et~al.(2013)Zemel, Wu, Swersky, Pitassi, and Dwork]{zemel2013learning}
Rich Zemel, Yu Wu, Kevin Swersky, Toni Pitassi, and Cynthia Dwork.
\newblock Learning fair representations.
\newblock In \emph{Proceedings of the 30th International Conference on Machine Learning}. PMLR, 2013.

\bibitem[Zhang et~al.(2024)Zhang, Fu, Fan, Zhang, Liu, Gu, Zhang, and Liu]{zhang2024compass}
Gaoyang Zhang, Bingtao Fu, Qingnan Fan, Qi Zhang, Runxing Liu, Hong Gu, Huaqi Zhang, and Xinguo Liu.
\newblock Compass: Enhancing spatial understanding in text-to-image diffusion models.
\newblock \emph{arXiv preprint arXiv:2412.13195}, 2024.

\bibitem[Zhang et~al.(2023)Zhang, Rao, and Agrawala]{zhang2023controlnet}
Lvmin Zhang, Anyi Rao, and Maneesh Agrawala.
\newblock Adding conditional control to text-to-image diffusion models.
\newblock In \emph{Proceedings of the IEEE/CVF international conference on computer vision}, pages 3836--3847, 2023.

\bibitem[Zhang et~al.(2025)Zhang, Tong, Zhao, Guo, Zhang, Zhang, Liu, Gao, and Li]{zhang2025let}
Renrui Zhang, Chengzhuo Tong, Zhizheng Zhao, Ziyu Guo, Haoquan Zhang, Manyuan Zhang, Jiaming Liu, Peng Gao, and Hongsheng Li.
\newblock Let's verify and reinforce image generation step by step.
\newblock In \emph{Proceedings of the Computer Vision and Pattern Recognition Conference}, pages 28662--28672, 2025.

\bibitem[Zhou et~al.(2024)Zhou, Li, Ma, Zhang, and Yang]{zhou2024migc}
Dewei Zhou, You Li, Fan Ma, Xiaoting Zhang, and Yi Yang.
\newblock Migc: Multi-instance generation controller for text-to-image synthesis.
\newblock In \emph{Proceedings of the IEEE/CVF conference on computer vision and pattern recognition}, pages 6818--6828, 2024.

\bibitem[Zhou et~al.(2025)Zhou, Shao, Bai, Zhang, Xu, Han, and Xie]{zhou2025golden}
Zikai Zhou, Shitong Shao, Lichen Bai, Shufei Zhang, Zhiqiang Xu, Bo Han, and Zeke Xie.
\newblock Golden noise for diffusion models: A learning framework.
\newblock In \emph{International Conference on Computer Vision}, 2025.

\end{thebibliography}
}

\maketitlesupplementary

\renewcommand{\thesection}{\Alph{section}}
\setcounter{section}{0}
\section{Comparison with Additional Methods}

\subsection{Comparison with COMPASS}
\label{supp_compass}

\revise{While our method demonstrates strong performance across various models and benchmarks, fine-tuning methods like COMPASS~\cite{zhang2024compass} can achieve higher accuracy in specific in-distribution settings, though often at the cost of degrading other alignment capabilities. For instance, on the GenEval benchmark with SD 1.4, COMPASS achieves a spatial accuracy of 43--46\% compared to our 36\% (Table~\ref{tab:benchmark_results}). We attribute this to the nature of fine-tuning, which modifies model weights and can lead to specialization on the training distribution. A more fine-grained analysis suggests this specialization may be narrow. On T2I-CompBench, which uses stricter geometric criteria, our method outperforms COMPASS for all models (e.g., 0.278 vs. 0.254 on SD 1.4); additionally, Table~\ref{tab:per_relation_accuracy} shows that COMPASS's advantage on SD 1.4 is largely driven by a significant performance gap on the ``left of'' relation (59.2\% vs. 40.8\%), while performance on other relations is more comparable.}

\revise{Importantly, COMPASS's specialization appears to come at a cost to generalization, particularly in complex, out-of-distribution scenarios (Tables \ref{tab:ood_results}, \ref{tab:multiple_relations}). This is most evident when handling prompts with multiple spatial relations—a compositional challenge not seen during training. As shown in Table~\ref{tab:multiple_relations}, our method's performance degrades gracefully, whereas COMPASS's accuracy drops sharply. For instance, on prompts with three objects and two relations involving OOD categories, our method achieves 28\% accuracy compared to COMPASS's 12\%. The performance gap widens with complexity: for four objects and three relations with OOD categories, our method maintains 19\% accuracy, while COMPASS drops to just 5\%. This highlights the advantage of our test-time optimization in maintaining compositional structure and degrading gracefully under complexity.}

\begin{figure}
  \centering
    \includegraphics[width=\linewidth]{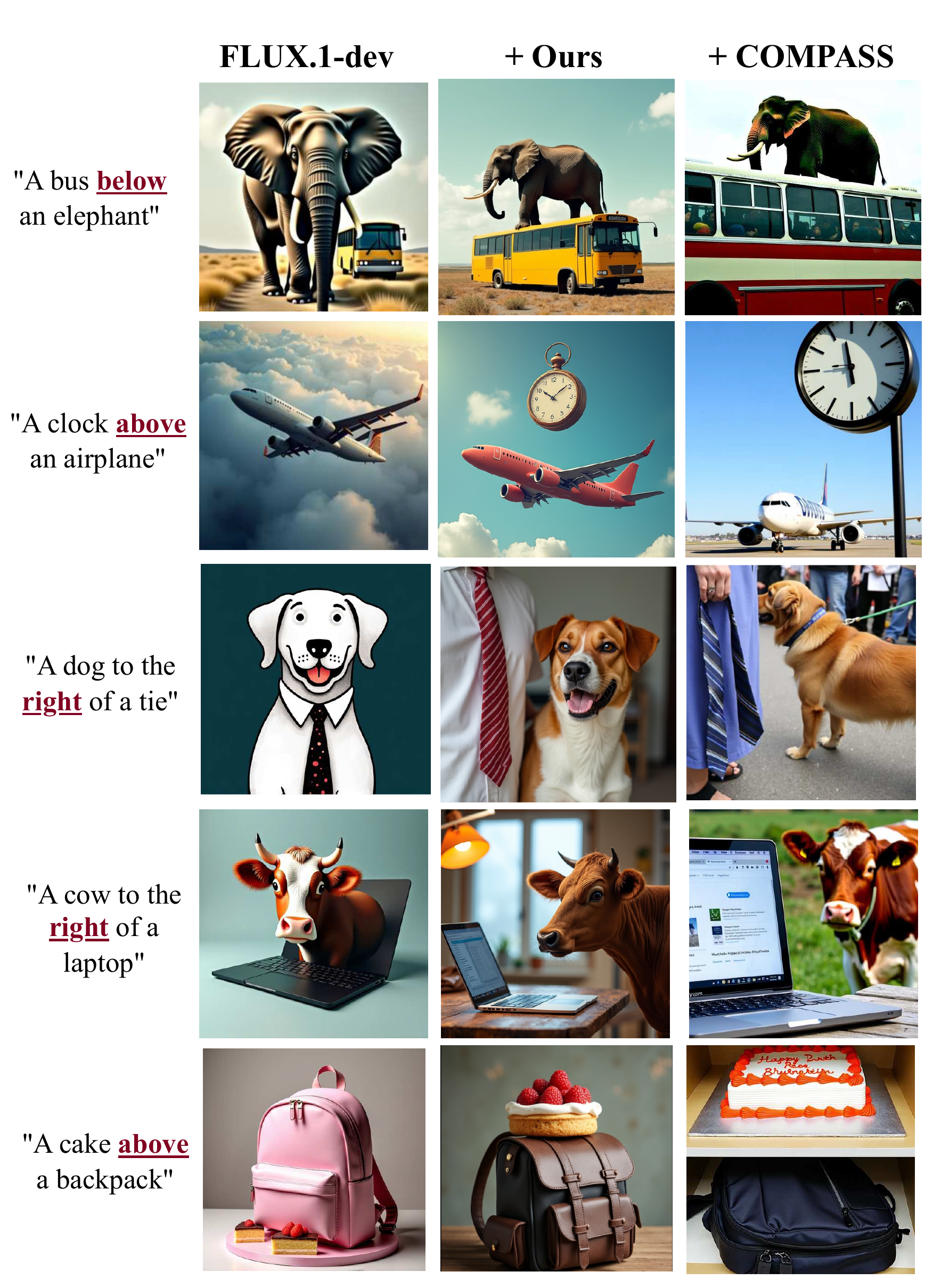}
   \caption{Comparing COMPASS~\cite{zhang2024compass} to our method, using FLUX.1-dev~\cite{flux} as the base model.}
   \label{fig:compass_qual}
\end{figure}

\begin{figure*}
  \centering
    \includegraphics[width=\linewidth]{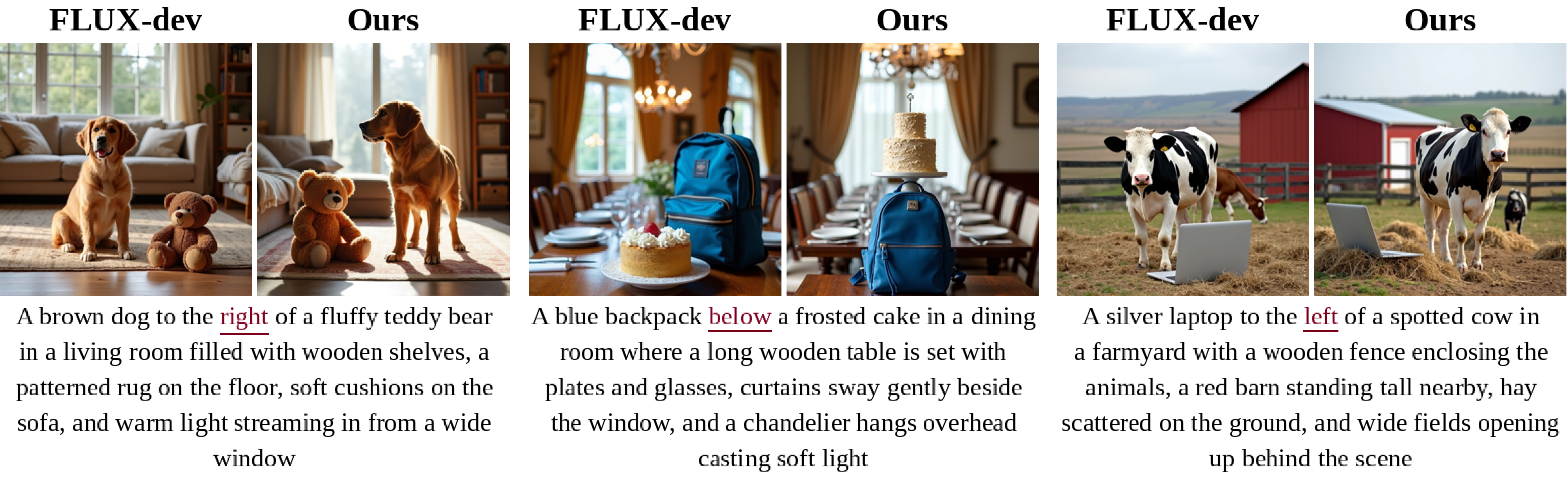}
   \caption{
   \revise{\textbf{Qualitative results with dense prompts.} Our method preserves the intended spatial relations even when objects are described with attributes and embedded in detailed scene contexts.}
   }
   \label{fig:dense_fig}
\end{figure*}

This trade-off extends to other capabilities beyond spatial reasoning. Our method better preserves the original aesthetics and style of the base model, as demonstrated in the qualitative comparison in~\cref{fig:compass_qual}. \revise{We quantify this observation in Table~\ref{tab:base_model_similarity}, which shows that images generated with our method have a higher visual similarity to the base model's generations using both DINO~\cite{oquab2023dinov2} and CLIP~\cite{radford2021learningCLIP} similarities (e.g., on FLUX.1-dev: 0.863 vs. 0.775 for DINO).} Furthermore, as shown in Table~\ref{tab:compass_tradeoffs}, COMPASS's improvements in spatial control are accompanied by degradation in other areas: on SD2.1, color accuracy drops from 0.85 to 0.71, and counting accuracy falls from 0.44 to 0.20. In contrast, our test-time approach is applied selectively only when spatial relations are present in the prompt, thereby preserving the base model's performance on these dimensions.

\begin{table}[ht]
  \centering
    \scalebox{0.95}
    {
    \renewcommand{\arraystretch}{0.9}
  \begin{tabular}{@{}llcc@{}}
    \toprule
    \revise{\textbf{Base Model}} & \revise{\textbf{Methods}} & \revise{\textbf{DINO~\textuparrow}} & \revise{\textbf{CLIP~\textuparrow}} \\
    
    \midrule
    \revise{\textbf{Flux-dev}} & \revise{} & \revise{} & \revise{} \\
    & \revise{Base vs. COMPASS} & \revise{0.7751} & \revise{0.7721} \\
    & \revise{Base vs. Ours} & \revise{\textbf{0.8625}} & \revise{\textbf{0.8767}} \\
    
    \midrule
    \revise{\textbf{SD 1.4}} & \revise{} & \revise{} & \revise{} \\
    & \revise{Base vs. COMPASS} & \revise{0.7135} & \revise{0.7530} \\
    & \revise{Base vs. Ours} & \revise{\textbf{0.7360}} & \revise{\textbf{0.8012}} \\
    
    \bottomrule
  \end{tabular}
  }
  \caption{\revise{Our approach generates images that are closer to the base model. In Figure~\ref{fig:compass_qual} we see that qualitatively style and aesthetics are closer to the base model. }}
  \label{tab:base_model_similarity}
\end{table}

\begin{table}[ht]
  \centering
    \scalebox{0.95}
    {
    \renewcommand{\arraystretch}{0.9}
  \begin{tabular}{@{}lcccc@{}}
    \toprule
    & \multicolumn{2}{c}{\revise{\textbf{FLUX.1-dev}}} & \multicolumn{2}{c}{\revise{\textbf{SD 1.4}}} \\
    \cmidrule(lr){2-3} \cmidrule(lr){4-5}
    \revise{\textbf{Relation}} & \revise{\textbf{Ours}} & \revise{\textbf{COMPASS}} & \revise{\textbf{Ours}} & \revise{\textbf{COMPASS}} \\
    
    \midrule
        \revise{above} & \revise{61.5\%} & \revise{63.5\%} & \revise{30.8\%} & \revise{39.4\%} \\
    \revise{below} & \revise{{67.0\%}} & \revise{54.5\%} & \revise{33.9\%} & \revise{38.4\%} \\
    \revise{left of} & \revise{60.5\%} & \revise{56.6\%} & \revise{40.8\%} & \revise{\textbf{59.2\%}} \\
    \revise{right of} & \revise{54.6\%} & \revise{{67.6\%}} & \revise{40.7\%} & \revise{40.7\%} \\
    
    \bottomrule
  \end{tabular}
  }
  \caption{\revise{Per-relation accuracy comparison. While FLUX.1-dev results are comparable between methods, in SD 1.4 COMPASS shows notably higher accuracy for ``left of" (59\%) compared to other relations.}}
  \label{tab:per_relation_accuracy}
\end{table}

\begin{table}[ht]
    \centering
    \scalebox{0.9}
    {
    \renewcommand{\arraystretch}{0.95}
    \begin{tabular}{@{}lcccc@{}}
        \toprule
        \textbf{Model} & \textbf{Position~\textuparrow} & \textbf{Count~\textuparrow} & \textbf{Color~\textuparrow} \\
        \midrule
        Vanilla SD2.1 & 0.07 & 0.44 & 0.85 \\
        COMPASS & \textbf{0.51} & 0.20 & 0.71 \\
        \ourmethod{} (Ours) & \textbf{0.54} & 0.44* & 0.85* \\
        \bottomrule
    \end{tabular}
    }
    \caption{GenEval results comparing fine-tuning (COMPASS) vs. test-time optimization (Ours) on SD2.1. COMPASS results are from their paper. *Our method preserves the vanilla model's performance on non-spatial tasks since we only apply steering when spatial relations are present in the prompt.}
    \label{tab:compass_tradeoffs}
\end{table}

\subsection{Comparison with Layout-guided Methods}
\label{supp_layout_guided}

\edit{Table~\ref{tab:layout_guided_results} compares \ourmethod{} with layout-guided approaches, such as RAGD~\cite{chen2025ragd}, 3DIS~\cite{zhou2025dis}, and MIGC~\cite{zhou2024migc}. Results for these methods are sourced from RAGD~\cite{chen2025ragd}, where GenEval's performance was evaluated using MLLM-generated layouts. Comparisons against MLLM-driven pipelines are inequitable due to their massive parameter counts and undisclosed training data. }

\begin{table}[ht]
  \centering
    \scalebox{0.9}
    {
    \renewcommand{\arraystretch}{0.95}
  \begin{tabular}{@{}lcc@{}}
    \toprule
    \textbf{Method} & \textbf{G. Pos.~\textuparrow} \\
    
    \midrule
    \edit{3DIS\dag~\cite{zhou2025dis}} & \edit{0.72} \\
    \edit{MIGC\dag~\cite{zhou2024migc}} & \edit{0.76} \\
    \edit{RAGD\dag~\cite{chen2025ragd}} & \edit{\textbf{0.80}} \\
    \edit{\ourmethod{} (Ours)} & \edit{0.61} \\
    
    \bottomrule
  \end{tabular}
  }
  \caption{
    \edit{Comparison of spatial relation scores from the GenEval benchmark (\textit{G. Pos.}) against layout-guided methods. \dag~ results are reported from~\cite{chen2025ragd}. Note that unlike \ourmethod{}, these methods rely on explicit layout inputs.}
  }
  \label{tab:layout_guided_results}
\end{table}

\section{Additional Ablation Studies}
\label{sec:supp_attn_agg_ablation}

\begin{table}[ht]
    \centering
    \scalebox{0.85}
    {
    \renewcommand{\arraystretch}{0.9}
    \begin{tabular}{llc}
        \toprule
        \textbf{Aggregation Variant} & \textbf{G. Pos.~\textuparrow} \\
        \midrule
        \edit{Full Concatenation (Ours)} & \edit{0.7} \\
        \edit{Mean: Per-Layer} & \edit{0.69} \\
        \edit{Mean: Global (All layers/heads)} & \edit{\textbf{0.78}} \\
        \bottomrule
    \end{tabular}
    }
    \caption{\edit{Ablation results for different attention aggregation strategies on FLUX.1-dev (GenEval validation set).}}
    \label{tab:attn_agg_ablation}
\end{table}

\begin{figure}[ht]
  \centering
    \includegraphics[width=\linewidth, keepaspectratio]
    {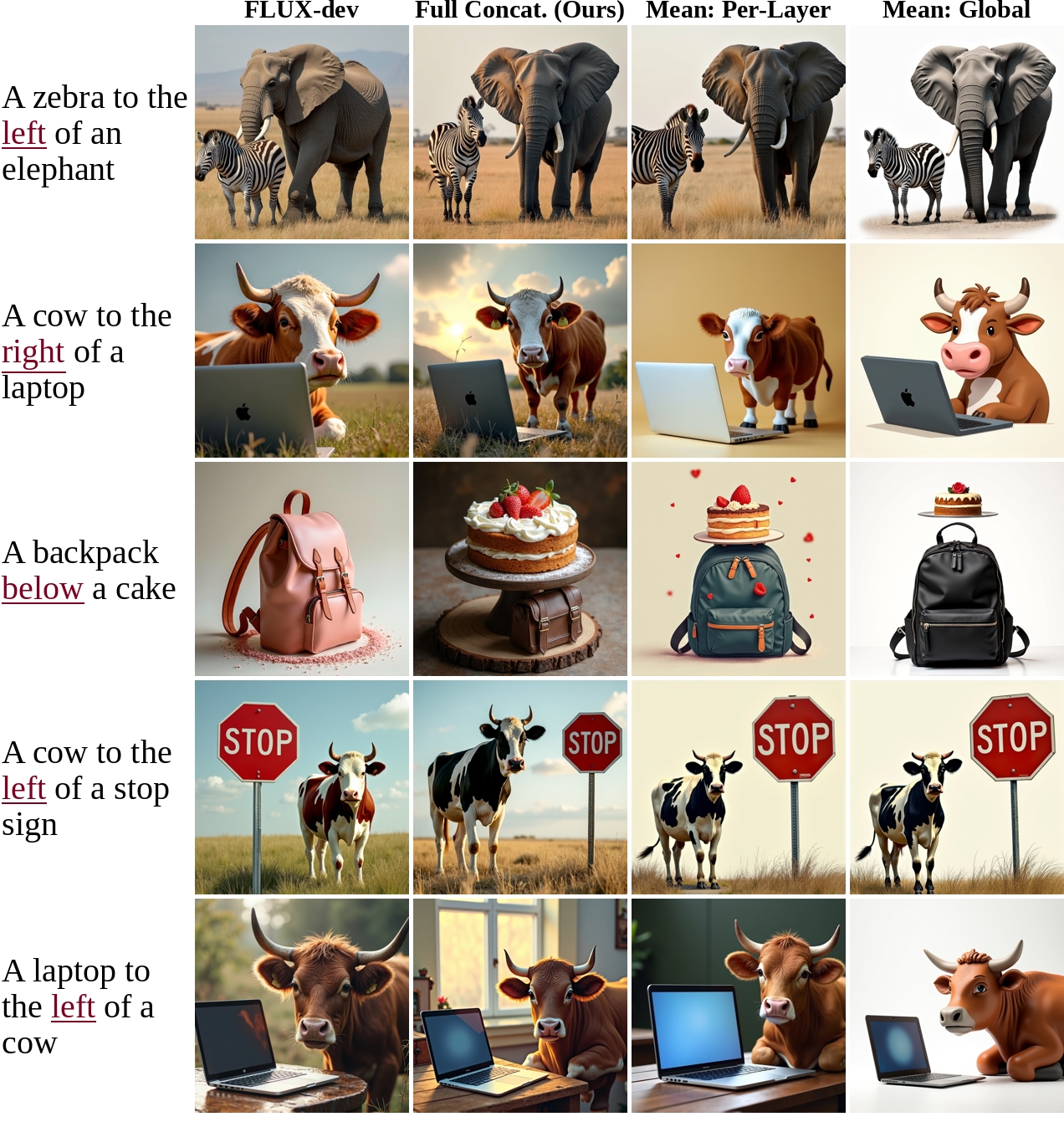} 
   \caption{\edit{\textbf{Impact of Aggregation on Visual Aesthetics.} Comparing our chosen method (Full Concatenation) against the Mean Aggregation baselines. While Mean Aggregation forces the objects into position, it frequently results in loss of background detail and vividness. Our method preserves the visual richness of the base model.}}
   \label{fig:attn_mean_poor_gen}
\end{figure}

\edit{
The inputs to the relation classifier are the cross-attention maps of two objects, alongside the current denoising timestep. We analyzed how different aggregation strategies for the attention maps affect performance. Specifically, we compared three approaches: \textbf{(1) Mean (Global):} Computing the mean across all layers and heads; \textbf{(2) Mean (Per-Layer):} Computing the mean across heads within each layer, then concatenating the layer outputs; and \textbf{(3) Full Concatenation:} Retaining all maps by concatenating all layers and heads into a single feature vector.
}

\edit{
 While Table~\ref{tab:attn_agg_ablation} demonstrates that the ``Mean (Global)'' strategy achieves the highest alignment score (0.78), we find that it comes at a cost to image quality, consistently yielding over-constrained results with ``dull'' or neutral backgrounds (See \cref{fig:attn_mean_poor_gen}). Similarly, the ``Mean (Per-Layer)'' strategy, which scores slightly lower than our method (0.69 vs. 0.70), also exhibits inconsistent visual quality, occasionally suffering from the same background degradation. In contrast, the ``Full Concatenation'' strategy provides the most balanced setup, offering strong adherence to the spatial prompt while maintaining high visual aesthetics. Therefore, we decided to use this strategy for our final method.
}

\section{Additional Implementation Details}
\label{supp_impl_details}

\subsection{Adapting FOR's handcrafted spatial loss to Flux-based models}
\label{supp_FOR_flux_impl}
We adapted FOR's~\cite{FOR_izadi2025fine} handcrafted spatial loss as follows: For each pair of objects, we first aggregated all cross-attention maps across all layers and heads into a single attention map per object. Subsequently, we applied the manual spatial loss function on the two aggregated maps in exactly the same manner as in the official implementation. To ensure a fair comparison, we ran a hyperparameter search over the scale factor of the step size used during test-time optimization to find the optimal setup.

\subsection{Benchmarks}
To ensure a fair comparison, we normalize all relation names to \spatialrelations{} across benchmarks. All baselines we reproduced use this normalized set. We will release normalized test sets for both benchmarks.

\subsection{{OOD Evaluation Details}}
\label{supp_ood_details}
\revise{Our out-of-distribution (OOD) evaluation tests generalization on 200 prompts across 16 object categories unseen by both our method and COMPASS~\cite{zhang2024compass}. We identified 6 categories from T2I-CompBench vocabulary that were out-of-distribution for both methods, and used ChatGPT to generate 10 additional diverse categories, ensuring no training data overlap.}

\subsection{{Extending GenEval to an Open-Set of Categories and Multiple Relations}}
\label{supp_multi_relation}

\revise{The current GenEval benchmark is constrained for only recognizing  the 80 COCO categories, and can only verify a single pair of relations.
To facilitate an open set of object categories, we replaced the object detector with CLIPSeg~\cite{clipseg_lueddecke22_cvpr}, a zero-shot object segmentation approach. To evaluate multiple relations, we generated metadata files in GenEval's expected structure, where each prompt is paired with a list of required objects along with their pairwise relations, verify each relation, and indicate success only when all relations are correctly generated in an image. To verify that our automatic evaluation is reliable, we compared its results against manual evaluation conducted by one of the authors on a mixed set of in-distribution and OOD categories (Table~\ref{tab:automatic_eval_mult_rels}). The scores align closely across all settings.
}

\begin{table}[ht]
  \centering
    \scalebox{0.95}
    {
    \renewcommand{\arraystretch}{0.9}
  \begin{tabular}{@{}ccccl@{}}
    \toprule
    \revise{\textbf{Objects}} & \revise{\textbf{Relations}} & \revise{\textbf{Manual Eval.}} & \revise{\textbf{Automatic Eval.}} \\
    \midrule
    \revise{3} & \revise{2} & \revise{0.29} & \revise{0.28} \\
    \revise{4} & \revise{3} & \revise{0.12} & \revise{0.12} \\
    \revise{5} & \revise{3} & \revise{0.07} & \revise{0.10} \\
    \bottomrule
  \end{tabular}
  }
  \caption{\revise{Comparison of automatic evaluation vs. manual evaluation of multiple spatial relations.}}
  \label{tab:automatic_eval_mult_rels}
\end{table}

\subsection{Object Categories for Multi-Relation Evaluation}
\label{supp_multi_relation_objects}
\revise{For the multi-relation evaluation, similar to the single-relation case, we tested on both in-distribution and out-of-distribution (OOD) object categories. The specific objects used were:}

\revise{\textbf{In-distribution categories (16 objects):} backpack, bowl, cat, skateboard, bottle, cake, tie, vase, sheep, toothbrush, teddy bear, suitcase, mouse, knife, chair, and umbrella.}

\revise{\textbf{Out-of-distribution categories (16 objects):} teapot, corgi, furby, robot, turtle, rabbit, butterfly, key, pig, bee, tiger, hammer, monkey, razor, snake, and stapler.}

\subsection{Extending Our Method to Diagonal Spatial Relations}
\label{diag_supp}

\revise{Since there is no labeled data for diagonal relations (top-left, top-right, bottom-left, bottom-right) readily available, we generated training samples by inferring relations directly from GQA bounding boxes. 
For each image, we examined all object pairs and computed their relative offsets and angles. Pairs whose center displacement fell within diagonal sectors were labeled accordingly, while cases with large overlaps were discarded.}

\revise{Using these labels, we constructed cross-attention training samples following the same procedure as for the four basic relations (above, below, left, right). For a given relation (e.g., \textit{``top-left''}), we inverted the image twice: once with the correct relation and once with an incorrect relation sampled uniformly from the remaining options (e.g., \textit{``bottom-left''}). This shows that our dual-image inversion strategy can naturally extend to more complex spatial relations such as diagonals.}

\revise{We then trained a diagonal relation classifier in the same way as for the four basic relations, with one exception: as a quick prototype of the aggregation module, we computed the average across all cross-attention maps before forwarding the representation to the classification module.}

\subsection{Potential limitations of dual-inversion}
\label{dual_inv_supp}
\revise{Our dual inversion strategy has an underlying assumption that the inversion process succeeds, and generates well-behaving attention maps even for the incorrect prompt. While we did not experiment with inverting very complex prompts, inversion it is likely to be less stable in these cases. 
Another limitation arises when inverting images that contain multiple instances of the same object (2 cats and a dog). Such images and prompts are known to be less stable.}

\begin{figure}
  \centering
    \includegraphics[width=0.7\linewidth]{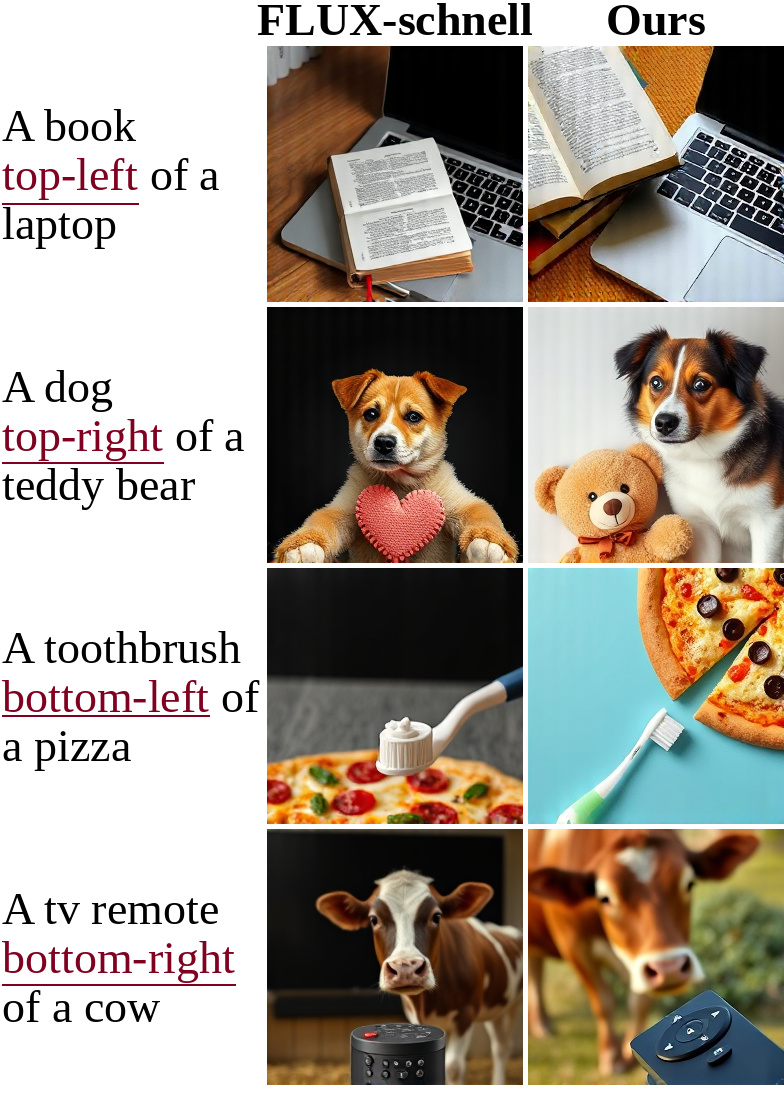}
   \caption{
   \revise{\textbf{Qualitative comparison of diagonal spatial relations.} The base model often misplaces objects or ignores the diagonal specification, while our method better adheres to the intended spatial layout.}
   }
   \label{fig:diag_qual_fig}
\end{figure}

\subsection{Dense Prompt Construction}
\label{dense_prompt_supp}
\revise{To create dense prompts, we asked ChatGPT~\cite{chatgpt} to enhance base prompts from the GenEval dataset, and generate an enriched variant. Each variant preserved the original spatial relation, added an attribute to every object, placed the objects in a realistic location (e.g., kitchen, living room, park), and included a vivid description of the location with detailed context.}

\subsection{Architecture Details}

\label{sec:architecture}

First we process each set of attention maps individually, using a transformer based aggregation module that maps each set of maps $L \times H \times h \times w$ to one single $h \times w$ map per image, where $L$, $H$ are the number of cross attention layers and heads and $h$, $w$ are the attention map dimensions. Then we process these two maps jointly to predict the relation class.

\noindent\textbf{Aggregation Module.} For a given object, the aggregator receives all $L \times H$ attention maps of size $h \times w$. Each map is treated as a separate token in a sequence of length $L \cdot H$. We then project each token from $h \cdot w$ dimensions to a shared embedding dimension $d=256$. This yields a sequence length $L \cdot H$ with a feature dimension of $d$, to which we add standard positional encodings \cite{vaswani2017attention} in $\mathbb{R}^{d}$ that signifies the spatial location. A timestep embedding $\phi(t) \in \mathbb{R}^{d}$ is also added to all tokens, in the input to every transformer layer, allowing the aggregator to adapt to the current denoising stage. A four-layer, four-head transformer then processes this sequence to exchange information across different layers and heads. To aggregate the transformer output from $\mathbb{R}^{L \cdot H, d}$ to $\mathbb{R}^{d}$, we use another four-layer, single-head transformer. 
For the second transformer, we learn a query vector in $\mathbb{R}^{1 \times d}$, while the key and value are projected from the output of the first transformer ($\mathbb{R}^{L \cdot H, d}$).  
Finally, a linear layer maps the output in $\mathbb{R}^{d}$ to $\mathbb{R}^{h\cdot w}$ reshaped to an aggregated map $\mathbf{A}_{\text{agg}}\in\mathbb{R}^{h\times w}$. The same aggregator (shared weights) processes both subject and object maps.

In practice, to process SD-based maps we use two-layer, two-head transformers instead of four-layer, four-head.

\noindent\textbf{Classification Module.}
The two aggregated maps are stacked into a two-channel image, flattened, projected to $d=256$ dimensions, tagged with standard positional encodings and passed through a four-layer eight-head Transformer. Finally, the transformer's output is concatenated with a low-dimension timestep embedding (16-d) and fed to an MLP that returns logits over the relation classes.
The complete architecture is trained jointly end-to-end.

In practice, to process SD-based maps we use two-layer, four-head transformers in the classification module.

\subsection{Other details}
\noindent\textbf{Relation Extraction.} Relation triplets are extracted from the prompt using SpaCy \citep{Honnibal_spaCy_Industrial-strength_Natural_2020} a natural lang. parsing library, with some hard-coded exceptions, like "fire hydrant".

\noindent\textbf{Classifier Training.} For training the classifier we used a batch size of 64, with AdamW optimizer \cite{loshchilov2018decoupled}, a weight decay of $0.05$, and a learning rate of $5\cdot10^{-5}$ for Flux-based models and $0.0001$ for SD-based models. We used a ReduceOnPlateu scheduler with a factor of $0.5$ and patience $5$. We trained the classifier for a maximum of $200$ epochs in Flux-based models and $150$ for SD-based models, with early stopping patience of $10$ epochs.

\noindent\textbf{Test-Time Optimization and Inference.}
We optimize the first $50\%$ of the denoising steps. In FLUX.1-schnell, we only optimize the initial noise. The step size $\alpha$ is $5$ for SD2.1 and Flux-based models, and $7.5$ for SD1.4 and number of optimization iterations per step are $15$.

We use the guidance scales suggested by HuggingFace \cite{huggingfaceT2I}, specifically $3.5$ for FLUX.1-dev, $0$ for FLUX.1-schnell, and $7.5$ for SD based models.

\noindent\textbf{Image Resolution.} We generate images at different resolutions depending on the base model: $256 \times 256$ for FLUX.1-schnell and $512 \times 512$ for FLUX.1-dev, SD2.1, and SD1.4.
\noindent\textbf{Neutral Relation Class.}
We included a fifth, "neutral" class during training. This class captures non-directional relationships (described by "is") and acts as a background category. We do not use this neutral class during generation.

\section{Additional Qualitative Results}
\revise{Figure~\ref{fig:unified_geneval} provides comprehensive comparisons across all four base models on GenEval prompts, demonstrating consistent improvements in spatial accuracy. Figure~\ref{fig_common_base_failures} \revise{(Appendix)}, demonstrate how our method overcomes the three main failure modes: incorrect object placement, entity neglect, and object fusion. }

\label{supp_qualitative_results}
We perform extensive comparisons on both GenEval (Figures~\ref{fig:schnell_geneval_r1}, \ref{fig:dev_geneval_r2}, \ref{fig:sd2_geneval_r2}, and \ref{fig:sd14_geneval_r2}) and T2I-CompBench (Figures~\ref{fig:dev_t2i_r1}, \ref{fig:schnell_t2i_r1}, \ref{fig:sd2_t2i_r1}, and \ref{fig:sd14_t2i_r1}).

\begin{figure*}
  \centering
    \includegraphics[width=\linewidth]{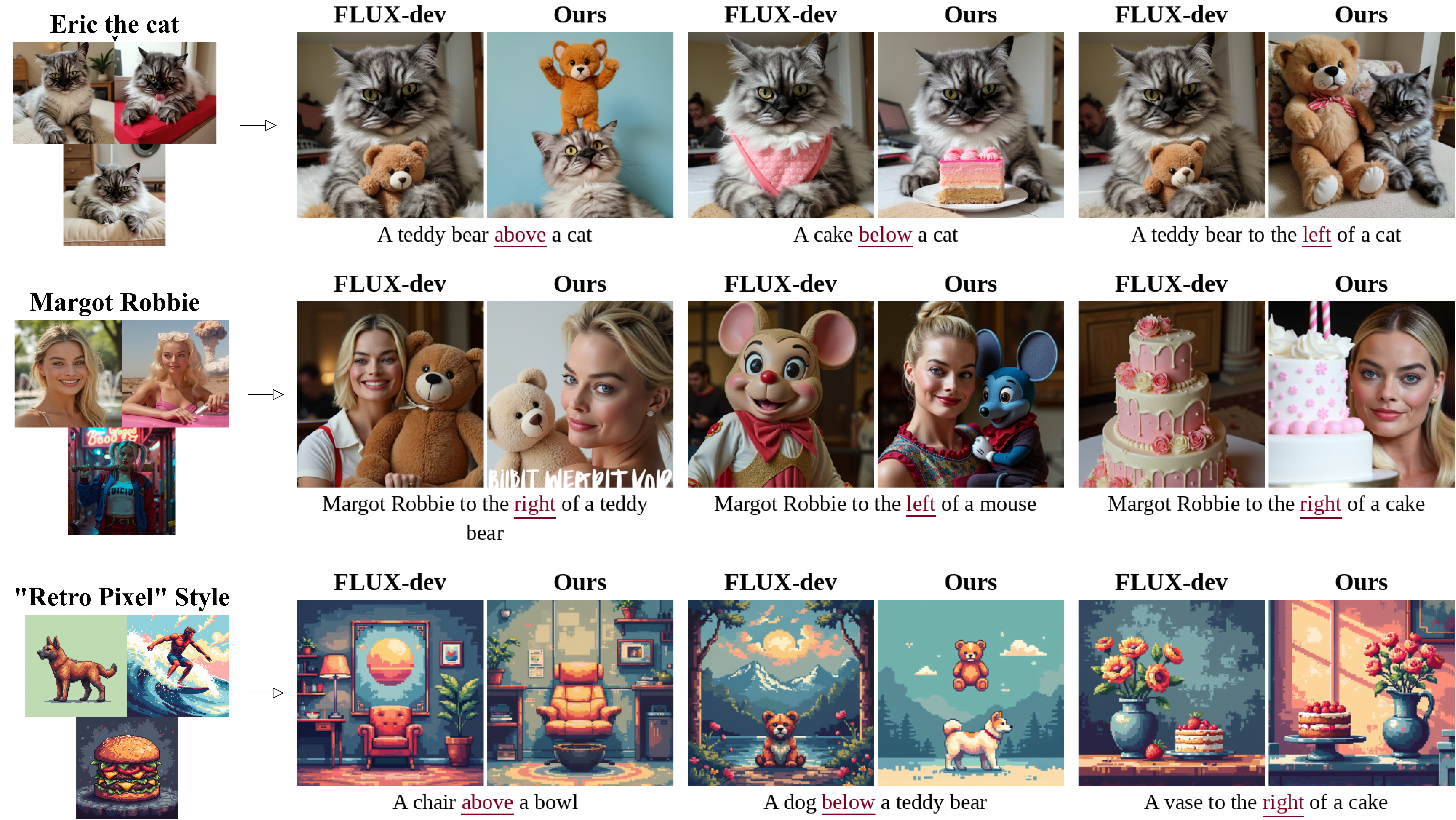}
    \caption{
    \revise{\textbf{More qualitative results of zero-shot transfer to personalized LoRAs of both subject and style.} }}%

   \label{fig:personal_fig_supp}
\end{figure*}

\begin{figure*}[htbp]
  \centering
    \includegraphics[width=0.95\linewidth, trim={0.cm 9.6cm 3.25cm 0cm},clip]{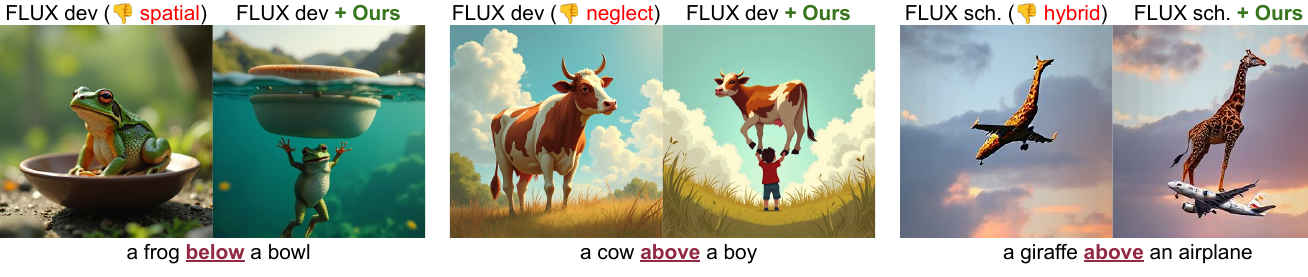} \\[-2pt]  %
\caption{\textbf{Common base model failures.} Our method addresses common text-to-image generation failures, including incorrect object placement (left), object neglect (middle), and fused, chimeric hybrids (right). }
   \label{fig_common_base_failures}
\end{figure*}

\begin{figure*}[htbp]
  \centering
    \includegraphics[width=\linewidth]{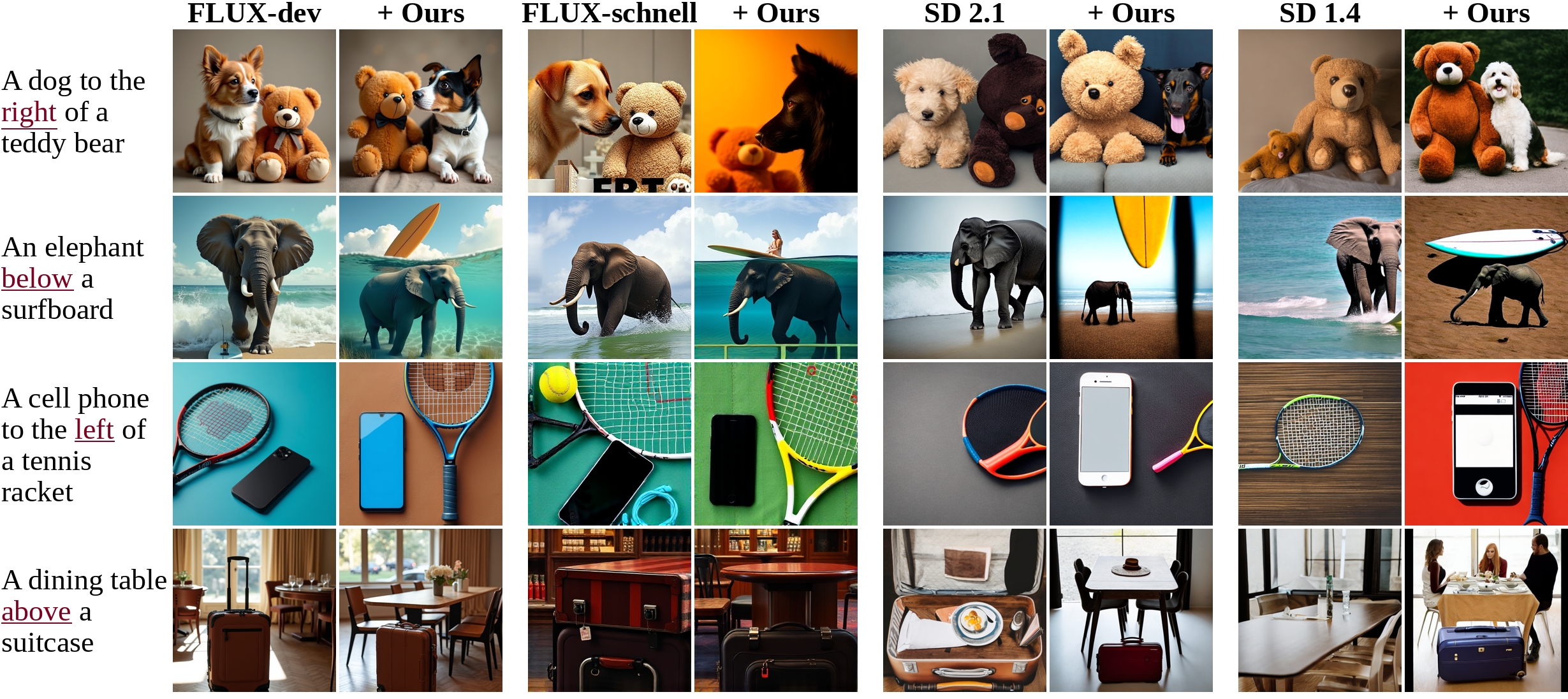}
   \caption{\textbf{Comparisons with all four base models}. Prompts were taken from  GenEval~\cite{ghosh2023geneval}. Each pair uses the same seed.} 
   \label{fig:unified_geneval}
\end{figure*}

\begin{figure*}
  \centering
    \includegraphics[width=\linewidth, height=1.20\linewidth, keepaspectratio]
    {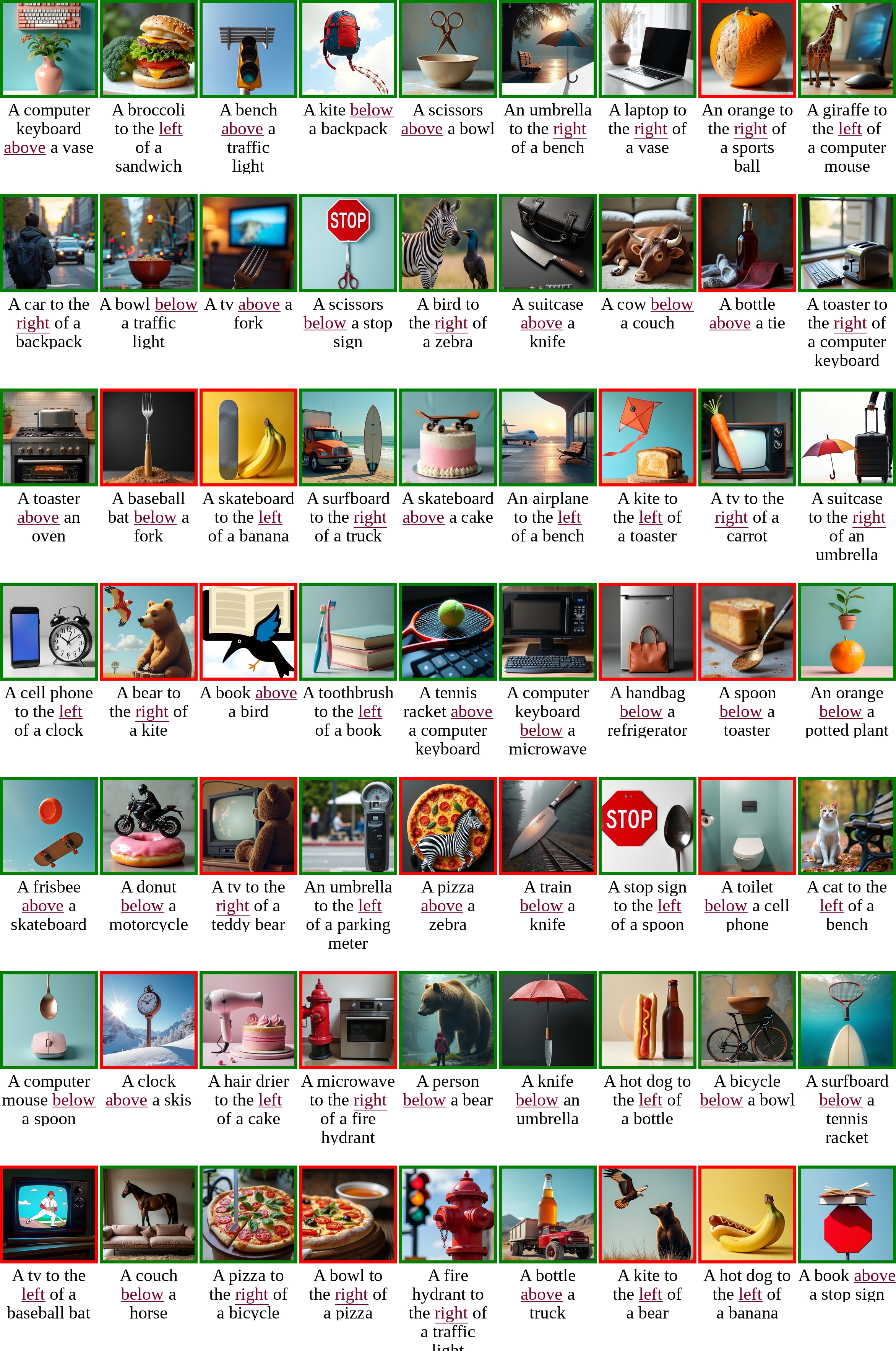}
   \caption{\textbf{Uncurated image generation results} from FLUX.1-dev~\cite{flux} with \ourmethod{}, using validation prompts from GenEval~\cite{ghosh2023geneval}. \revise{Green boxes mark spatially-aligned images (per GenEval), while red boxes mark misaligned ones.}}
   \label{fig:uncurated}
\end{figure*}

\begin{figure}
  \centering
    \includegraphics[width=\linewidth, height=2.5\linewidth, keepaspectratio]
    {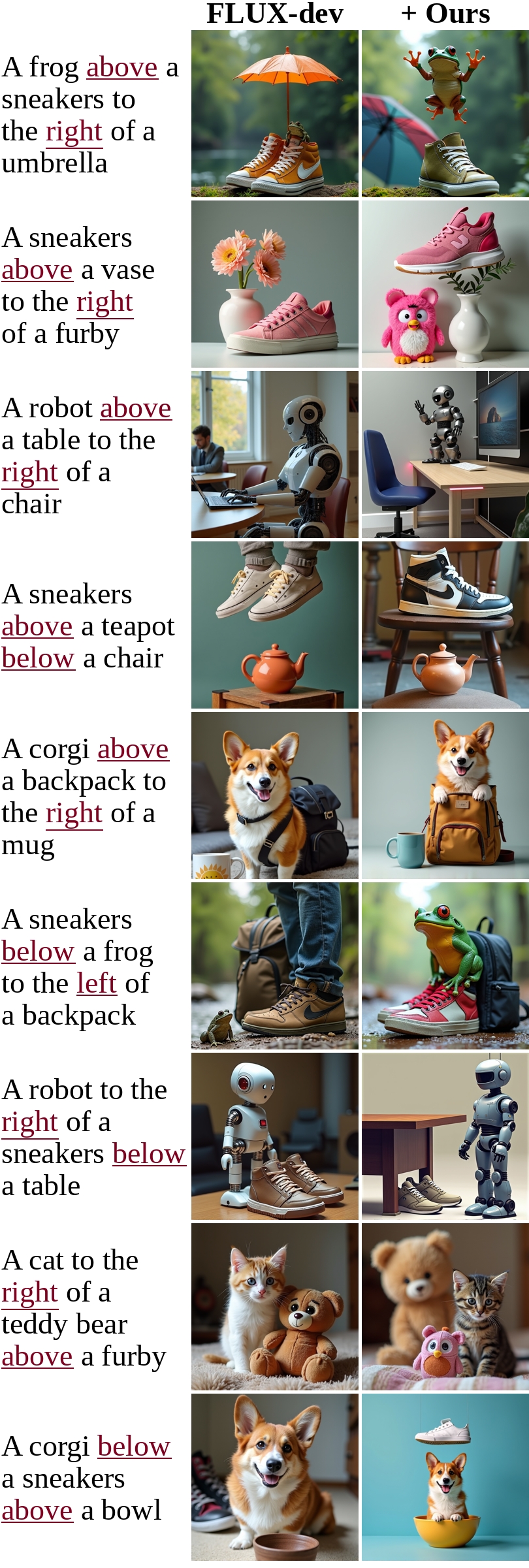}
   \caption{\textbf{Handling Multiple Relations.} As opposed to the base model, our method is capable of generating complex scene structures that contain multiple relations in a single prompt.}
   \label{fig:supp_multi_rel_qual}
\end{figure}

\begin{figure}
  \centering
    \includegraphics[width=\linewidth, height=2.5\linewidth, keepaspectratio]
    {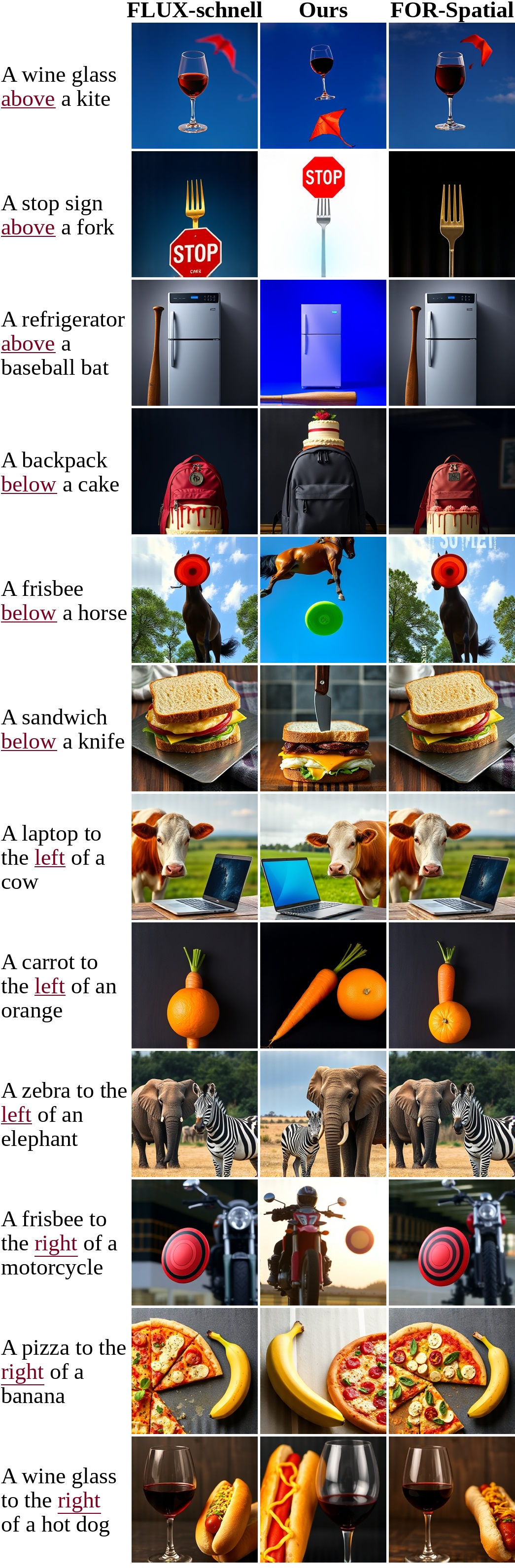}

   \caption{Qualitative comparison using FLUX.1-schnell~\cite{flux} with prompts from the GenEval~\cite{ghosh2023geneval} benchmark. For each prompt, the same seed is used for all methods.}
   \label{fig:schnell_geneval_r1}
\end{figure}

\begin{figure}
  \centering
    \includegraphics[width=\linewidth, keepaspectratio]
    {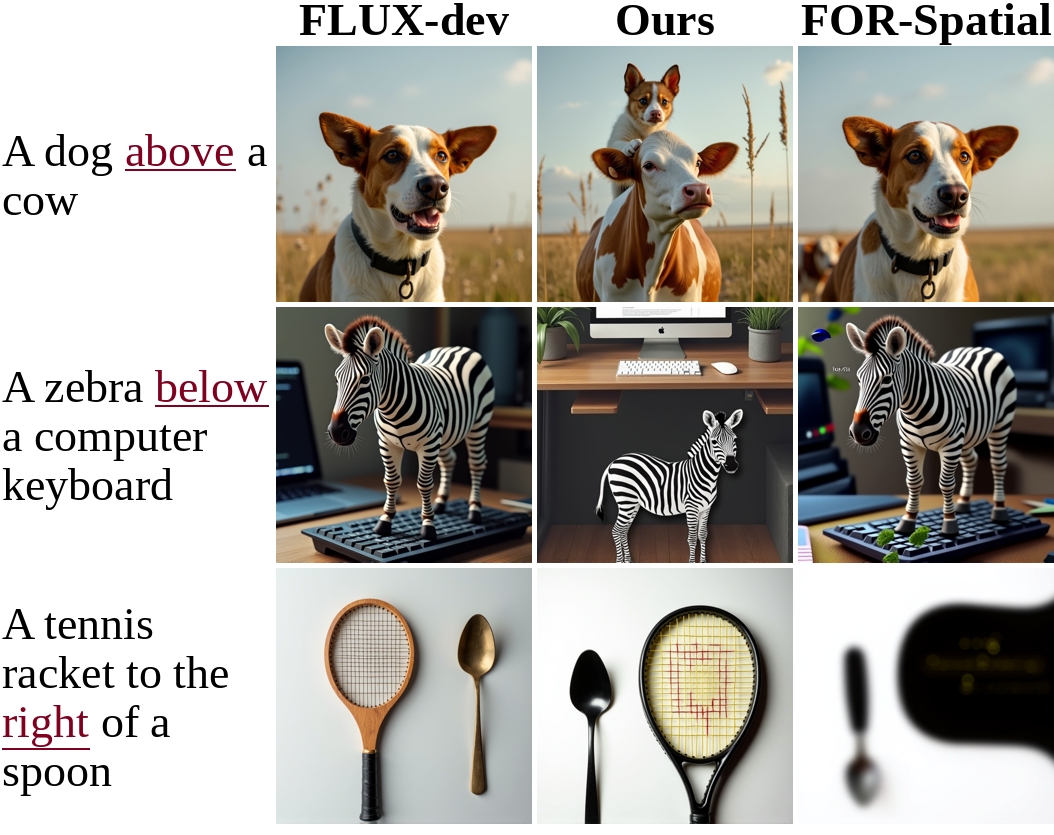}
   \caption{FLUX.1-dev on the GenEval benchmark}
   \label{fig:dev_geneval_r2}
\end{figure}

\begin{figure*}
  \centering
    \includegraphics[width=\linewidth]
    {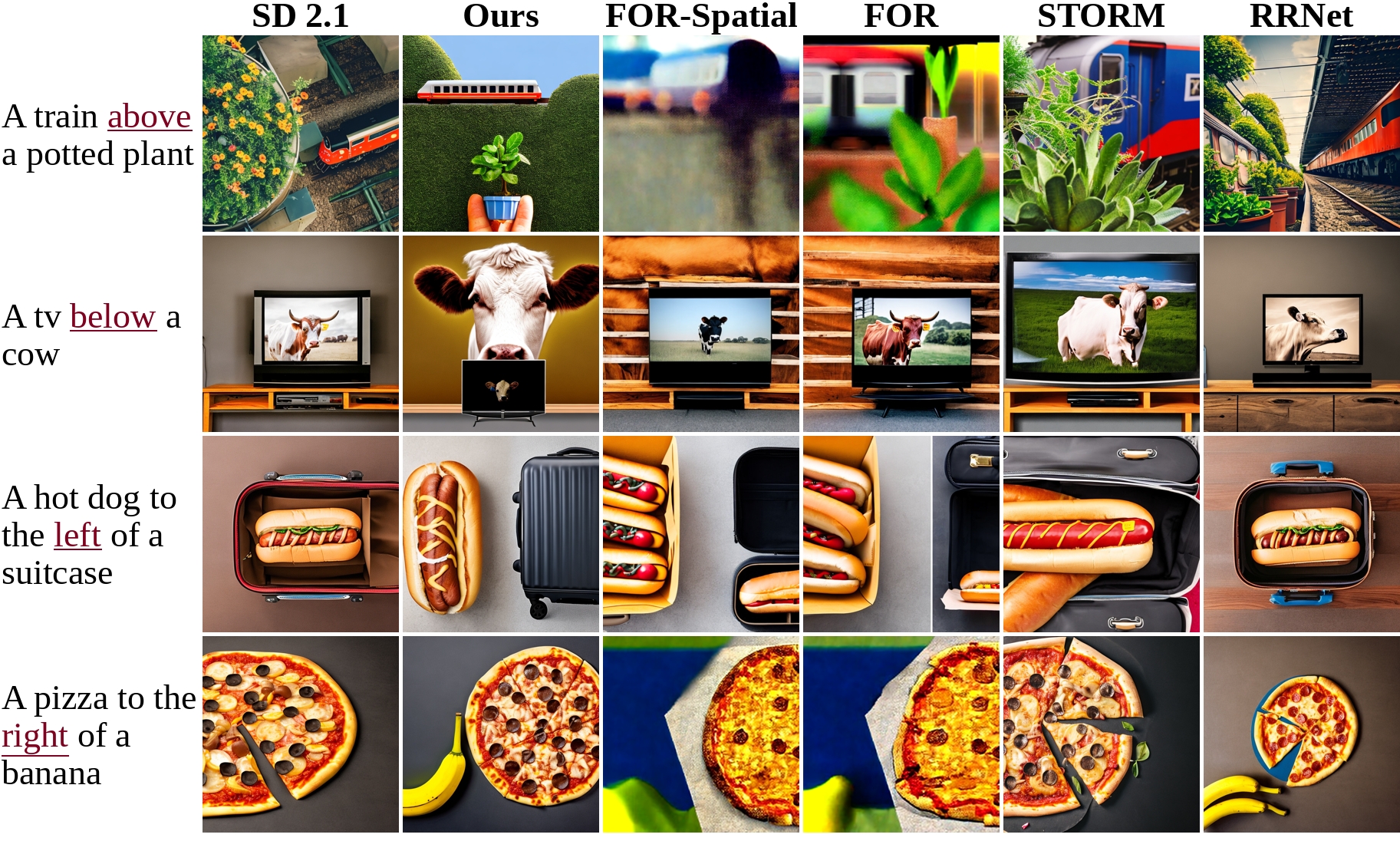}
   \caption{SD 2.1 on the GenEval benchmark}
   \label{fig:sd2_geneval_r2}
\end{figure*}

\begin{figure*}
  \centering
    \includegraphics[width=\linewidth, height=1.2\linewidth, keepaspectratio]
    {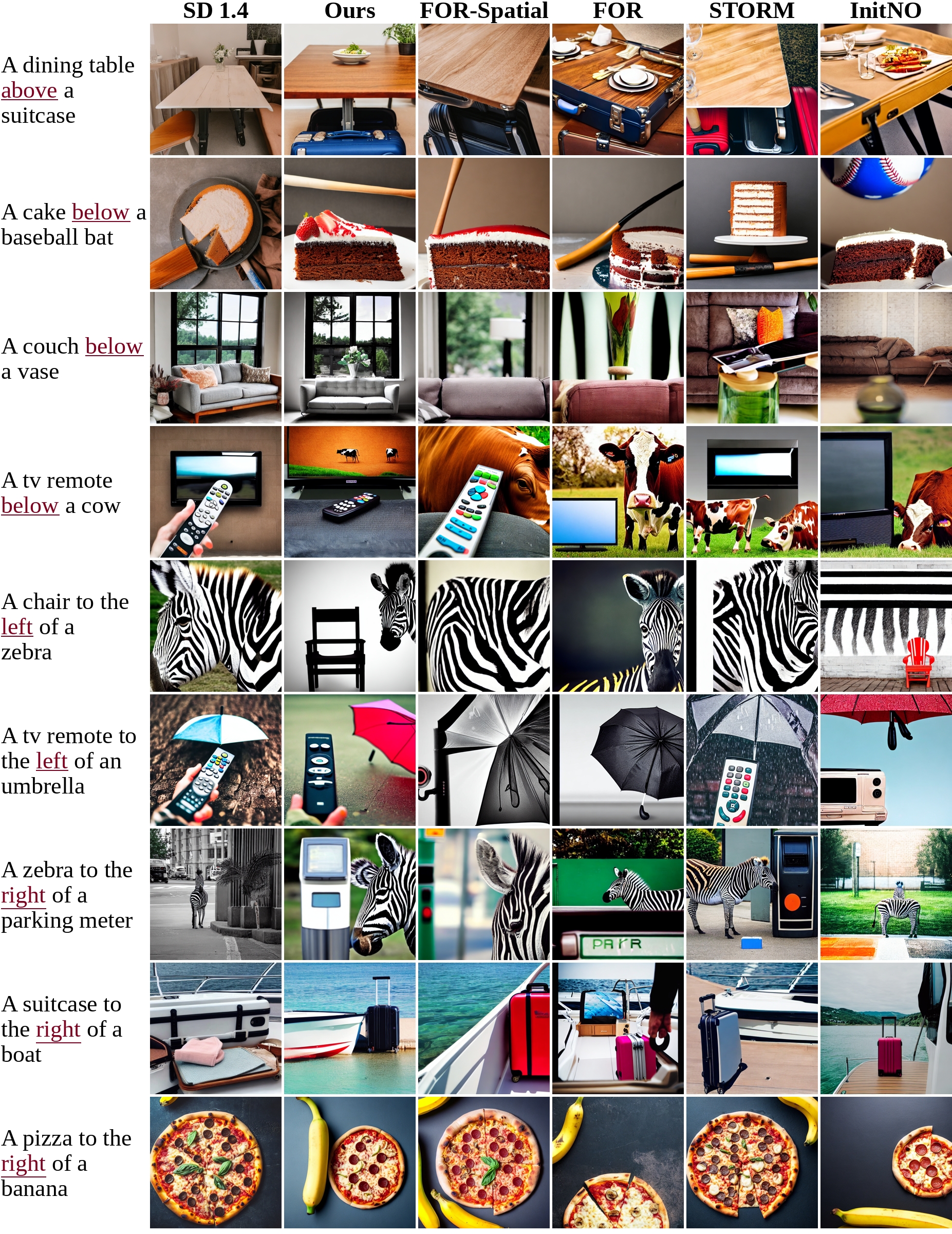}
   \caption{SD 1.4 on the GenEval benchmark}
   \label{fig:sd14_geneval_r2}
\end{figure*}

\begin{figure}
  \centering
    \includegraphics[width=\linewidth, height=2.5\linewidth, keepaspectratio]
    {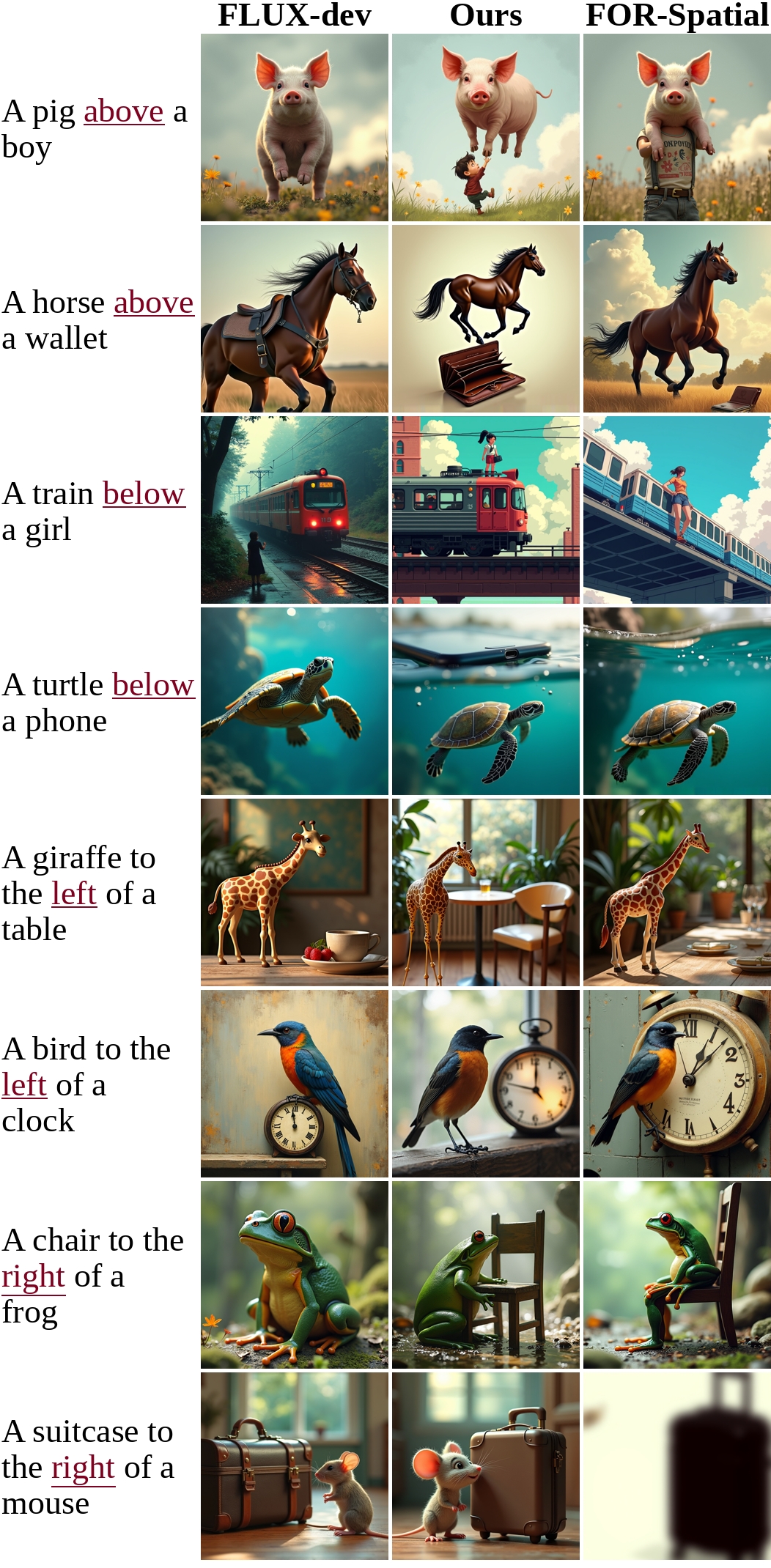}
   \caption{FLUX.1-dev on the T2I-CompBech benchmark}
   \label{fig:dev_t2i_r1}
\end{figure}

\begin{figure}
  \centering
    \includegraphics[width=\linewidth, height=2.5\linewidth, keepaspectratio]
    {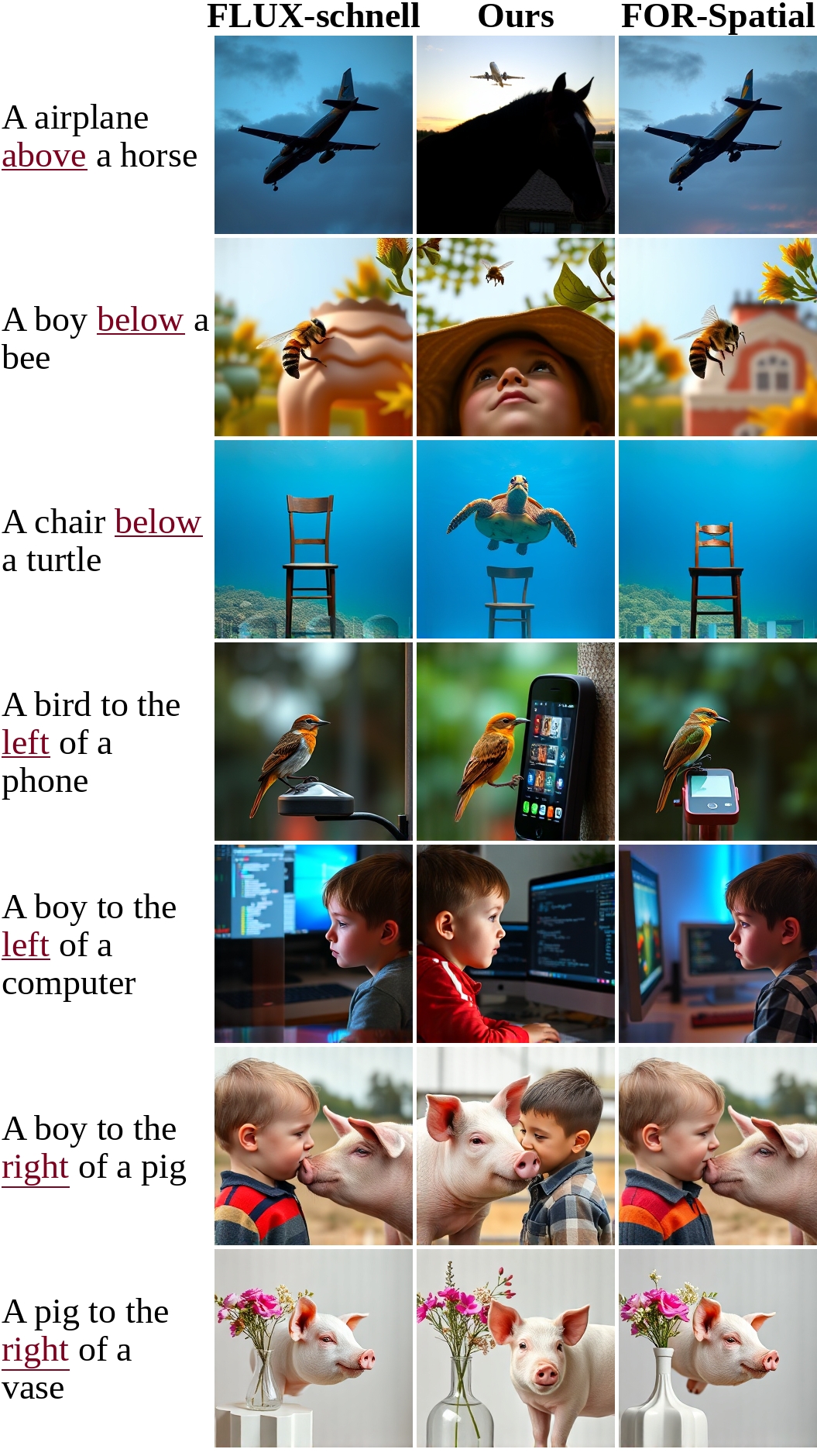}
   \caption{FLUX.1-schnell on the T2I-CompBech benchmark}
   \label{fig:schnell_t2i_r1}
\end{figure}

\begin{figure*}
  \centering
    \includegraphics[width=\linewidth]
    {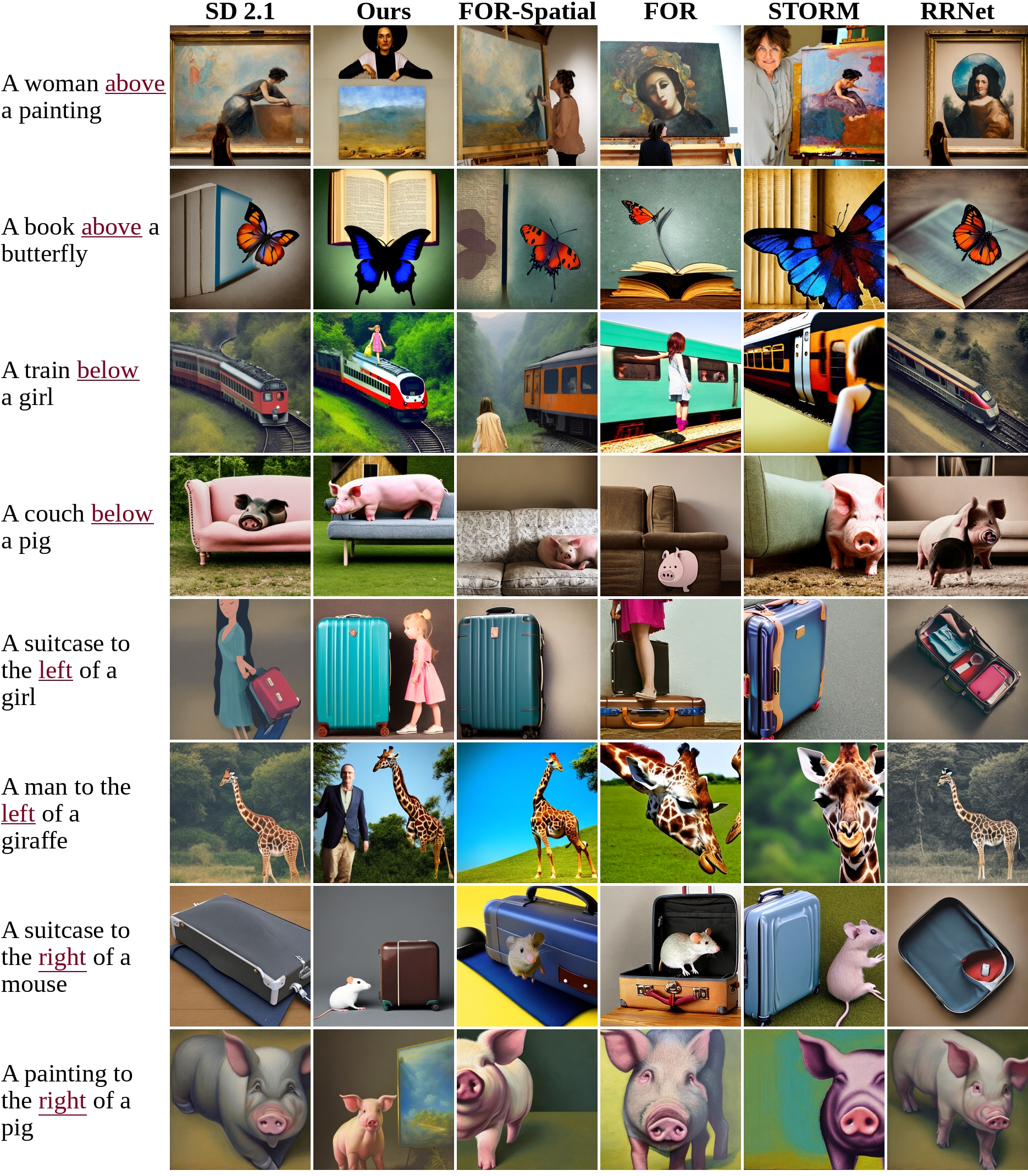}
   \caption{SD 2.1 on the T2I-CompBech benchmark}
   \label{fig:sd2_t2i_r1}
\end{figure*}

\begin{figure*}
  \centering
    \includegraphics[width=\linewidth]
    {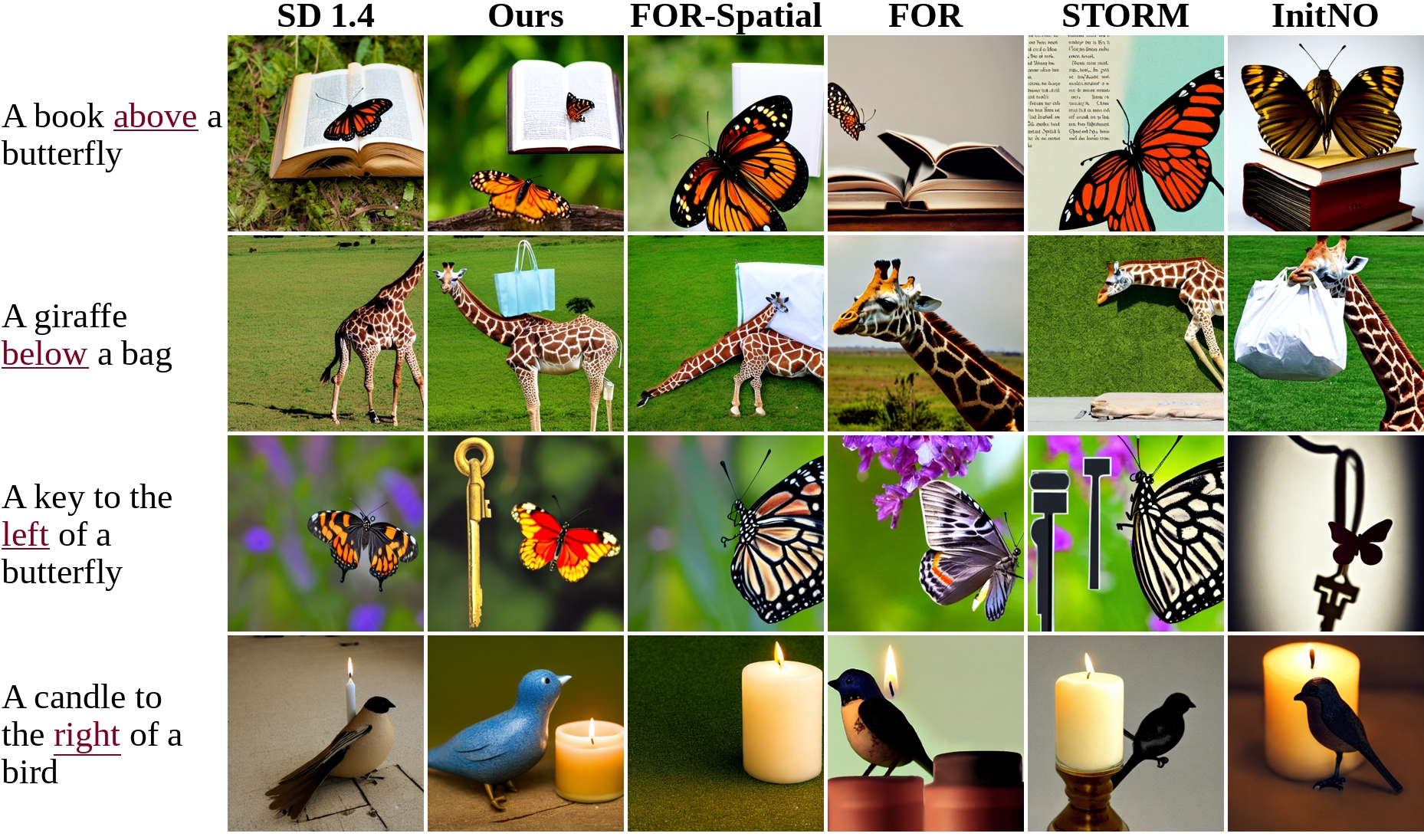}
   \caption{SD 1.4 on the T2I-CompBech benchmark}
   \label{fig:sd14_t2i_r1}
\end{figure*}

\end{document}